\documentclass[12pt]{article}

\newtheorem{theorem}{Theorem}[section]
\newtheorem{definition}{Definition}[section]
\newtheorem{lemma}{Lemma}[section]

\newtheorem{remark}{Remark}[section]

\usepackage{natbib}
\usepackage{graphicx}
\usepackage{amssymb,amsmath}

\oddsidemargin
-0.2in \topmargin -.60in
\begin{document}

\title{Generalization Bounds for Domain Adaptation}

\author{
Chao~Zhang$^{1}$,\quad Lei~Zhang$^{2}$,\quad Jieping Ye$^{1,3}$\\
$^1$Center for Evolutionary
Medicine and Informatics, The Biodesign Institute, \\and $^3$Computer Science and Engineering,
Arizona State University,
Tempe, USA\\
\texttt{zhangchao1015@gmail.com; jieping.ye@asu.edu} \\
$^{2}$School of Computer Science and Technology,\\
Nanjing University of Science and Technology,
Nanjing, P.R. China\\
\texttt{zhanglei.njust@yahoo.com.cn} \\
}



\maketitle

\begin{abstract}
In this paper, we provide a new framework to obtain the generalization bounds of the learning process for domain adaptation, and then apply the derived bounds to analyze the asymptotical convergence of the learning process. Without loss of generality, we consider two kinds of representative domain adaptation: one is with multiple sources and the other is combining source and target data.\\
\indent In particular, we use the integral probability metric to measure the difference between two domains. For either kind of domain adaptation, we develop a related Hoeffding-type deviation inequality and a symmetrization inequality to achieve the corresponding generalization bound based on the uniform entropy number. We also generalized the classical McDiarmid's inequality to a more general setting where independent random variables can take values from different domains. By using this inequality, we then obtain generalization bounds based on the Rademacher complexity. Afterwards, we analyze the asymptotic convergence and the rate of convergence of the learning process for such kind of domain adaptation. Meanwhile, we discuss the factors that affect the asymptotic behavior of the learning process and the numerical experiments support our theoretical findings as well.
\end{abstract}


\section{Introduction}

The generalization bound measures the probability that a function, chosen from a function class by an algorithm, has a sufficiently small error and plays an important role in statistical learning theory \citep[see][]{Vapnik99,Bousquet04}. The generalization bounds have been widely used to study the consistency of the ERM-based learning process \citep{Vapnik99}, the asymptotic convergence of empirical process \citep{Vaart96} and the learnability of learning models \citep{Blumer89}. Generally, there are three essential aspects to obtain the generalization bounds of a specific learning process: complexity measures of function classes, deviation (or concentration) inequalities and symmetrization inequalities related to the learning process. For example, \citet{Vaart96} presented the generalization bounds based on the Rademacher complexity and the covering number, respectively. \citet{Vapnik99} gave the generalization bounds based on the Vapnik-Chervonenkis (VC) dimension. \citet{Bartlett05} proposed the local Rademacher complexity and obtained a sharp generalization bound for a particular function class $\{f\in\mathcal{F}:\mathrm{E}f^2<\beta\mathrm{E}f, \beta>0\}$. \citet{Hussain11} showed improved loss bounds for multiple kernel learning.

It is noteworthy that the aforementioned results of statistical learning theory are all built under the assumption that training and test data are drawn from the same distribution (or briefly called the assumption of same distribution). This assumption may not be valid in the situation that training and test data have different distributions, which will arise in many practical applications including speech recognition \citep{Jiang07} and natural language processing \citep{Blitzer07a}. Domain adaptation has recently been proposed to handle this situation and it is aimed to apply a learning model, trained by using the samples drawn from a certain domain ({\it source domain}), to the samples drawn from another domain ({\it target domain}) with a different distribution \citep[see][]{Bickel07,Wu04,Blitzer06,Ben-David10,Bian12}.

Without loss of generality, this paper is mainly concerned with two types of representative domain adaptation. In the first type, the learner receives training data from several source domains, known as {\it domain adaptation with multiple sources} \citep[see][]{Crammer06,Crammer08,Mansour08,Mansour09}. In the second type, the learner minimizes a convex combination of the source and the target empirical risks, termed as {\it domain adaptation combining source and target data} \citep[see][]{Ben-David10,Blitzer08}.

\subsection{Overview of Main Results}


In this paper, we present a new framework to obtain the generalization bounds of the learning process for the aforementioned two kinds of representative domain adaptation, respectively. Based on the resultant bounds, we then analyze the asymptotical properties of the learning processes for the two types of domain adaptation. There are four major aspects in the framework:
\begin{itemize}
  \item the quantity measuring the difference between two domains;
\item the complexity measure of function class;
\item the deviation inequalities of the learning process for domain adaptation;
\item the symmetrization inequality of the learning precess for domain adaptation.
\end{itemize}

Generally, in order to obtain the generalization bounds of a learning process, one needs to develop the related deviation (or concentration) inequalities of the learning process.
 For either kind of domain adaptation, we use a martingale method to develop the related Hoeffding-type deviation inequality. Moreover, in the situation of domain adaptation, since the source domain differs from the target domain, the desired symmetrization inequality for domain adaptation should incorporate some quantity to reflect the difference. From the point of this view, we then obtain the related symmetrization inequality incorporating the integral probability metric that measures the difference between the distributions of the source and the target domains.
Next, we present the generalization bounds based on the uniform entropy number for both kinds of domain adaptation. Also, we generalize the classical McDiarmid's inequality to a more general setting, where independent random variables take values from different domains. By using the derived inequality, we obtain the generalization bounds based on the Rademacher complexity.
 Following the resultant bounds, we study the asymptotic convergence and the rate of convergence of the learning process in addition to a discussion on factors that affect the asymptotic behaviors. The numerical experiments support our theoretical findings as well. Meanwhile, we give a comparison with the related results under the assumption of same distribution.

\subsection{Organization of the Paper}

The rest of this paper is organized as follows. Section \ref{sec:notation-setup} introduces the problems studied in this paper. Section \ref{sec:Distance} introduces the integral probability metric to measure the difference between two domains. In Section \ref{sec:UEN}, we introduce two kinds of complexity measures of function classes including the uniform entropy number and the Rademacher complexity.
In Section \ref{sec:MS} (resp. Section \ref{sec:CSTD}), we present the generalization bounds of the learning process for domain adaptation with multiple sources (resp. combining source and target data), and then analyze the asymptotic behavior of the learning process in addition to the related numerical experiment supporting our findings. In Section \ref{sec:compare}, we list the existing works on the theoretical analysis of domain adaptation as a comparison and the last section concludes the paper. In the appendices, we prove main results of this paper. For clarity of presentation, we also postpone the discussion of the deviation inequalities and the symmetrization inequalities in the appendices.


\section{Problem Setup} \label{sec:notation-setup}

In this section, we formalize the main issues of this paper by introducing some necessary notations

\subsection{Domain Adaptation with Multiple Sources}

We denote $\mathcal{Z}^{(S_k)}:=\mathcal{X}^{(S_k)}\times\mathcal{Y}^{(S_k)}\subset\mathbb{R}^I\times\mathbb{R}^J$ $(1\leq k\leq K)$ and $\mathcal{Z}^{(T)}:=\mathcal{X}^{(T)}\times\mathcal{Y}^{(T)}\subset\mathbb{R}^I\times\mathbb{R}^J$ as the $k$-th source domain and the target domain, respectively. Set $L=I+J$.
Let $\mathcal{D}^{(S_k)}$ and $\mathcal{D}^{(T)}$ stand for the distributions of the input spaces $\mathcal{X}^{(S_k)}$ $(1\leq k\leq K)$ and $\mathcal{X}^{(T)}$, respectively. Denote $g^{(S_k)}_*:\mathcal{X}^{(S_k)}\rightarrow\mathcal{Y}^{(S_k)}$ and $g^{(T)}_*:\mathcal{X}^{(T)}\rightarrow\mathcal{Y}^{(T)}$ as the labeling functions of $\mathcal{Z}^{(S_k)}$ ($1\leq k\leq K$) and $\mathcal{Z}^{(T)}$, respectively. In the situation of domain adaptation with multiple sources, the input-space distributions  $\mathcal{D}^{(S_k)}$ $(1\leq k\leq K)$ and $\mathcal{D}^{(T)}$ differ from each other, or $g^{(S_k)}_*$ $(1\leq k\leq K)$ and $g^{(T)}_*$ differ from each other, or both of the cases occur. There are sufficient amounts of i.i.d. samples ${\bf Z}_{1}^{N_k}=\{{\bf z}^{(k)}_n\}_{n=1}^{N_k}$ drawn from each source domain $\mathcal{Z}^{(S_k)}$ $(1\leq k\leq K)$ but little or no labeled samples drawn from the target domain $\mathcal{Z}^{(T)}$.



Given ${\bf w}=(w_1,\cdots,w_K)\in[0,1]^{K}$ with $\sum_{k=1}^Kw_k=1$, let $g_{\bf w}\in\mathcal{G}$ be the function
that minimizes the empirical risk
\begin{equation}\label{eq:emrisk.L}
\mathrm{E}^{(S)}_{\bf w}(\ell\circ g)=\sum_{k=1}^Kw_k\mathrm{E}_{N_k}^{(S_k)}(\ell\circ g)=\sum_{k=1}^K\frac{w_k}{N_k}\sum_{n=1}^{N_k}\ell(g({\bf x}^{(k)}_n),{\bf
    y}^{(k)}_n)
 \end{equation}
 over $\mathcal{G}$ with respect to sample sets $\{{\bf Z}_{1}^{N_k}\}_{k=1}^K$, and it is expected that $g_{\bf w}$ will perform well on the target expected risk:
\begin{equation}\label{eq:t.exrisk}
    \mathrm{E}^{(T)}(\ell\circ g):=\int\ell(g({\bf x}^{(T)}),{\bf y}^{(T)})d\mathrm{P}({\bf z}^{(T)}),\;\;g\in\mathcal{G},
\end{equation}
{\it i.e.,} $g_{\bf w}$ approximates the labeling $g^{(T)}_*$ as precisely as possible.

In the learning process of domain adaptation with multiple sources, we are mainly interested in the following two types of quantities:
\begin{itemize}
\item $\mathrm{E}^{(T)}(\ell\circ g_{\bf w})-\mathrm{E}^{(S)}_{\bf w}(\ell\circ g_{\bf w})$, which corresponds to the estimation of the expected risk in the target domain $\mathcal{Z}^{(T)}$ from the empirical quantity that is the weighted combination of the empirical risks in the multiple sources $\{\mathcal{Z}^{(S_k)}\}_{k=1}^K$;
\item $\mathrm{E}^{(T)}(\ell\circ g_{\bf w})-\mathrm{E}^{(T)}(\ell\circ \widetilde{g}_*)$, which corresponds to the performance of the algorithm for domain adaptation with multiple sources,
\end{itemize}
where $\widetilde{g}_*\in\mathcal{G}$ is the function that minimizes the expected risk $\mathrm{E}^{(T)}(\ell\circ g)$ over $\mathcal{G}$.

Recalling \eqref{eq:emrisk.L} and \eqref{eq:t.exrisk}, since
\begin{equation*}
   \mathrm{E}^{(S)}_{\bf w}(\ell\circ \widetilde{g}_*)-\mathrm{E}^{(S)}_{\bf w}(\ell\circ g_{\bf w})\geq 0,
\end{equation*}
we have
\begin{align}\label{eq:induction.M}
 \mathrm{E}^{(T)}(\ell\circ g_{\bf w})=&\mathrm{E}^{(T)}(\ell\circ g_{\bf w})-\mathrm{E}^{(T)}(\ell\circ \widetilde{g}_*)+\mathrm{E}^{(T)}(\ell\circ \widetilde{g}_*)\nonumber\\
\leq&\mathrm{E}^{(S)}_{\bf w}(\ell\circ \widetilde{g}_*)-\mathrm{E}^{(S)}_{\bf w}(\ell\circ g_{\bf w})+\mathrm{E}^{(T)}(\ell\circ g_{\bf w})-\mathrm{E}^T(\ell\circ \widetilde{g}_*)+\mathrm{E}^T(\ell\circ \widetilde{g}_*)\nonumber\\
\leq& 2\sup_{g\in\mathcal{G}}\big|\mathrm{E}^{(T)}(\ell\circ g)-\mathrm{E}_{\bf w}^{(S)}(\ell\circ g)\big|  +\mathrm{E}^{(T)}(\ell\circ \widetilde{g}_*),
\end{align}
and thus
\begin{eqnarray*}
   0 \leq \mathrm{E}^{(T)}(\ell\circ g_{\bf w})-\mathrm{E}^{(T)}(\ell\circ \widetilde{g}_*)
   \leq 2\sup_{g\in\mathcal{G}}\big|\mathrm{E}^{(T)}(\ell\circ g)-\mathrm{E}_{\bf w}^{(S)}(\ell\circ g)\big|.
\end{eqnarray*}
This shows that the asymptotic behaviors of the aforementioned two quantities when the sample numbers $N_1,\cdots,N_K$ go to {\it infinity} can both be described by the supremum
\begin{equation}\label{eq:upbound1}
\sup_{g\in\mathcal{G}}\big|\mathrm{E}^{(T)}(\ell\circ g)-\mathrm{E}_{\bf w}^{(S)}(\ell\circ g)\big|,
\end{equation}
which is the so-called generalization bound of the learning process for domain adaptation with multiple sources.

For convenience, we define the loss function as
class
\begin{equation}\label{eq:fuclass}
    \mathcal{F}:=\{{\bf z} \mapsto \ell(g({\bf x}),{\bf y}):g\in \mathcal{G}\},
\end{equation}
and call $\mathcal{F}$ as the function class in the rest of this paper.
By \eqref{eq:emrisk.L} and \eqref{eq:t.exrisk}, given sample sets $\{{\bf Z}_{1}^{N_k}\}_{k=1}^K$ drawn from  $\{\mathcal{Z}^{(S_k)}\}_{k=1}^K$ respectively, we briefly denote for any $f\in\mathcal{F}$,
\begin{equation}\label{eq:short1.M}
    \mathrm{E}^{(T)}f:=\int f({\bf z}^{(T)})d\mathrm{P}({\bf z}^{(T)})\;,
\end{equation}
and
\begin{equation}\label{eq:short2.M}
   \mathrm{E}^{(S)}_{\bf w}f:=\sum_{k=1}^K\frac{w_k}{N_k}\sum_{n=1}^{N_k}f({\bf z}^{(k)}_n).
\end{equation}
Thus, we rewrite the generalization bound \eqref{eq:upbound1} for domain adaptation with multiple sources as
\begin{equation}\label{eq:upbound2}
\sup_{f\in\mathcal{F}}\big|\mathrm{E}^{(T)}f-\mathrm{E}_{\bf w}^{(S)}f\big|.
\end{equation}


\subsection{Domain Adaptation Combining Source and Target Data}

Denote $\mathcal{Z}^{(S)}:=\mathcal{X}^{(S)}\times\mathcal{Y}^{(S)}\subset\mathbb{R}^I\times\mathbb{R}^J$ and $\mathcal{Z}^{(T)}:=\mathcal{X}^{(T)}\times\mathcal{Y}^{(T)}\subset\mathbb{R}^I\times\mathbb{R}^J$ as the source domain and the target domain, respectively. Let $\mathcal{D}^{(S)}$ and $\mathcal{D}^{(T)}$ stand for the distributions of the input spaces $\mathcal{X}^{(S)}$ and $\mathcal{X}^{(T)}$, respectively. Denote $g^{(S)}_*:\mathcal{X}^{(S)}\rightarrow\mathcal{Y}^{(S)}$ and $g^{(T)}_*:\mathcal{X}^{(T)}\rightarrow\mathcal{Y}^{(T)}$ as the labeling functions of $\mathcal{Z}^{(S)}$ and $\mathcal{Z}^{(T)}$, respectively. In the situation of domain adaptation combining source and target data \citep[see][]{Blitzer08,Ben-David10}, the input-space distributions $\mathcal{D}^{(S)}$ and $\mathcal{D}^{(T)}$ differ from each other, or the labeling functions $g^{(S)}_*$ and $g^{(T)}_*$ differ from each other, or both cases occur. There are some (but not enough) samples ${\bf Z}_{1}^{N_{T}}:=\{{\bf z}_n^{(T)}\}_{n=1}^{N_T}$ drawn from the target domain $\mathcal{Z}^{(T)}$ in addition to a large amount of samples ${\bf Z}_{1}^{N_{S}}:=\{{\bf z}_n^{(S)}\}_{n=1}^{N_S}$ drawn from the source domain $\mathcal{Z}^{(S)}$ with $N^{(T)}\ll N^{(S)}$. Given a $\tau\in[0,1)$, we denote $g_{\tau}\in\mathcal{G}$ as the function that minimizes the convex combination of the source and the target empirical risks over $\mathcal{G}$:
\begin{equation}\label{eq:error.C}
\mathrm{E}_{\tau}(\ell\circ g):=\tau\mathrm{E}^{(T)}_{N_T}(\ell\circ g)+(1-\tau)\mathrm{E}^{(S)}_{N_S}(\ell\circ g),
\end{equation}
and it is expected that $g_{\tau}$ will perform well for any pair ${\bf z}^{(T)}=({\bf x}^{(T)},{\bf y}^{(T)})\in\mathcal{Z}^{(T)}$, {\it i.e.}, $g_{\tau}$ approximates the labeling function $g^{(T)}_*$ as precisely as possible.

As mentioned by \citet{Blitzer08,Ben-David10}, setting $\tau$ involves a tradeoff between the source data that are sufficient but not accurate and the target data that are accurate but not sufficient. Especially, setting $\tau=0$ provides a learning process of the basic domain adaptation with one single source \citep[see][]{Ben-David06}.

Similar to the situation of domain adaptation with multiple sources, two types of quantities: $\mathrm{E}^{(T)}(\ell\circ g_{\tau})-\mathrm{E}_\tau (\ell\circ g_{\tau})$ and $\mathrm{E}^{(T)}(\ell\circ g_{\tau})-\mathrm{E}^{(T)}(\ell\circ \widetilde{g}_*)$ also play an essential role in analyzing the asymptotic behavior of the learning process for domain adaptation combining source and target data.
By the similar way of \eqref{eq:induction.M}, we need to consider the supremum
\begin{equation}\label{eq:upbound2.C}
\sup_{g\in\mathcal{G}}\big|\mathrm{E}^{(T)}(\ell\circ g)
-\mathrm{E}_{\tau}(\ell\circ g)\big|,\vspace{-1mm}
\end{equation}
which is the so-called generalization bound of the learning process for domain adaptation combining source and target data.
Following the notation of \eqref{eq:fuclass} and taking $f=\ell\circ g$, we can equivalently rewrite the generalization bound \eqref{eq:upbound2.C} as
\begin{equation}\label{eq:upbound3.C}
\sup_{f\in\mathcal{F}}\big|\mathrm{E}^{(T)}f-\mathrm{E}_{\tau}f\big|.
\end{equation}


\section{Integral Probability Metric}\label{sec:Distance}

As shown in some existing works \citep[see][]{Mansour08,Mansour09,Ben-David10,Ben-David06}, one of major challenges in the theoretical analysis of domain adaptation is to find a quantity to measure the difference between the source domain $\mathcal{Z}^{(S)}$ and the target domain $\mathcal{Z}^{(T)}$. Then, one can use the quantity to achieve generalization bounds for domain adaptation. In this section, we use the integral probability metric to measure the difference between the distributions of $\mathcal{Z}^{(S)}$ and $\mathcal{Z}^{(T)}$, and then discuss the relationship between the integral probability metric and other quantities proposed in existing works, {\it e.g.}, the {\it $\mathcal{H}$-divergence} and the {\it discrepancy distance} \citep[see][]{Ben-David10,Mansour09b}. Moreover, we will show that there is a special situation of domain adaptation, where the integral probability metric performs better than other quantities (see Remark \ref{rem:metric})

\subsection{Integral Probability Metric}

In \citet{Ben-David10,Ben-David06}, the {\it $\mathcal{H}$-divergence} was introduced to derive the generalization bounds based on the VC dimension under the condition of ``$\lambda$-close". \citet{Mansour09b} obtained the generalization bounds based on the Rademacher complexity by using the {\it discrepancy distance}. Both quantities are
aimed to measure the difference between two input-space distributions $\mathcal{D}^{(S)}$ and $\mathcal{D}^{(T)}$. Moreover, \citet{Mansour09} used the R\'enyi divergence to measure the distance between two distributions. In this paper, we use the following quantity to measure the difference between the distributions of the
source and the target domains:
\begin{definition}\label{def:distance}
Given two domains $\mathcal{Z}^{(S)},\mathcal{Z}^{(T)}\subset\mathbb{R}^L$, let ${\bf z}^{(S)}$ and ${\bf z}^{(T)}$ be the random variables taking values from $\mathcal{Z}^{(S)}$ and $\mathcal{Z}^{(T)}$, respectively. Let $\mathcal{F}\subset\mathbb{R}^{\mathcal{Z}}$ be a function class.
We define
\begin{equation}\label{eq:distance}
D_{\mathcal{F}}(S,T):=\sup_{f\in\mathcal{F}}|\mathrm{E}^{(S)}f-\mathrm{E}^{(T)}f|,
\end{equation}
where the expectations $\mathrm{E}^{(S)}$ and $\mathrm{E}^{(T)}$ are taken on the distributions $\mathcal{Z}^{(S)}$ and $\mathcal{Z}^{(T)}$, respectively.
\end{definition}

The quantity $D_{\mathcal{F}}(S,T)$ is termed as the integral probability metric that has played an important role in probability theory for measuring the difference between the two probability distributions \citep[see][]{Zolotarev84,Rachev,Muller97,Reid11}. Recently, \citet{Sriperumbudur09,Sriperumbudur12} gave the further investigation and proposed an empirical method to compute the integral probability metric. As mentioned by \citet{Muller97}[page 432],
the quantity $D_{\mathcal{F}}(S,T)$ is a semimetric and it is a metric if and only if the function class $\mathcal{F}$ separates the set of all signed measures with $\mu(\mathcal{Z}) = 0$. Namely, according to Definition \ref{def:distance}, given a non-trivial function class $\mathcal{F}$, the integral probability metric $D_{\mathcal{F}}(S,T)$ is equal to {\it zero} if the domains $\mathcal{Z}^{(S)}$ and $\mathcal{Z}^{(T)}$ have the same distribution.

By \eqref{eq:fuclass}, the quantity $D_{\mathcal{F}}(S,T)$ can be equivalently rewritten as
\begin{align}\label{eq:distance2}
D_{\mathcal{F}}(S,T)=&\sup_{g\in\mathcal{G}}\Big|\mathrm{E}^{(S)}\ell(g({\bf x}^{(S)}),{\bf y}^{(S)})-\mathrm{E}^{(T)}\ell(g({\bf x}^{(T)}),{\bf y}^{(T)})\Big|\nonumber\\
=&\sup_{g\in\mathcal{G}}\Big|\mathrm{E}^{(S)}\ell\big(g({\bf x}^{(S)}),g_*^{(S)}({\bf x}^{(S)})\big)-\mathrm{E}^{(T)}\ell\big(g({\bf x}^{(T)}),g_*^{(T)}({\bf x}^{(T)})\big)\Big|.
\end{align}
Next, based on the equivalent form \eqref{eq:distance2}, we discuss the relationships between the quantity $D_{\mathcal{F}}(S,T)$ and other quantities including the {\it $\mathcal{H}$-divergence} and the {\it discrepancy distance}.

\subsection{Relationship with Other Quantities}

Before the formal discussion, we briefly introduce the related quantities proposed in existing works \citep[see][]{Ben-David10,Mansour09b}.

\subsubsection{$\mathcal{H}$-Divergence and Discrepancy Distance}

In classification tasks, by setting $\ell$ as the absolute-value loss function ($\ell({\bf x},{\bf y})=|{\bf x}-{\bf y}|$), \citet{Ben-David10} introduced a variant of the {\it $\mathcal{H}$-divergence}:
\begin{align}\label{eq:H-div}
   d_{\mathcal{H}\triangle\mathcal{H}}(\mathcal{D}^{(S)},\mathcal{D}^{(T)})
=\sup_{g_1,g_2\in\mathcal{H}}\Big|\mathrm{E}^{(S)}\ell\big(g_1({\bf x}^{(S)}),g_2({\bf x}^{(S)})\big)-\mathrm{E}^{(T)}\ell\big(g_1({\bf x}^{(T)}),g_2({\bf x}^{(T)})\big)\Big|
\end{align}
to achieve VC-dimension-based generalization bounds for domain adaptation under the condition of ``$\lambda$-close": there exists a $\lambda>0$ such that
\begin{align*}
    \lambda\geq
\inf_{g\in\mathcal{G}}\left\{\int\ell(g({\bf x}^{(S)}),{\bf y}^{(S)})d\mathrm{P}({\bf z}^{(S)})+\int\ell(g({\bf x}^{(T)}),{\bf y}^{(T)})d\mathrm{P}({\bf z}^{(T)})\right\}.
\end{align*}

In both of the classification and regression tasks, given a function class $\mathcal{G}$ and a loss function $\ell$, \citet{Mansour09b} defined the {\it discrepancy distance} as
\begin{align}\label{eq:d-distance}
\mathrm{disc}_{\ell}(\mathcal{D}^{(S)},\mathcal{D}^{(T)})=\sup_{g_1,g_2\in\mathcal{G}}\Big|\mathrm{E}^{(S)}\ell\big(g_1({\bf x}^{(S)}),g_2({\bf x}^{(S)})\big)-\mathrm{E}^{(T)}\ell\big(g_1({\bf x}^{(T)}),g_2({\bf x}^{(T)})\big)\Big|,
\end{align}
and then used this quantity to obtain the generalization bounds based on the Rademacher complexity.

As mentioned by \citet{Mansour09b}, the quantities \eqref{eq:H-div} and \eqref{eq:d-distance} match in the setting of classification tasks by setting $\ell$ as the absolute-value loss function, while the usage of \eqref{eq:d-distance} does not require the condition of ``$\lambda$-close" but the usage of \eqref{eq:H-div} does. Recalling Definition \ref{def:distance}, since there is no limitation on the function class $\mathcal{F}$, the integral probability metric
$D_{\mathcal{F}}(S,T)$ can be used in both classification and regression tasks.
Therefore, we only consider the relationship between the integral probability metric $D_{\mathcal{F}}(S,T)$ and the {\it discrepancy distance} $\mathrm{disc}_{\ell}(\mathcal{D}^{(S)},\mathcal{D}^{(T)})$.

\subsubsection{Relationship between $D_{\mathcal{F}}(S,T)$ and $\mathrm{disc}_{\ell}(\mathcal{D}^{(S)},\mathcal{D}^{(T)})$}

From Definition \ref{def:distance} and \eqref{eq:distance2}, we can find that the integral probability metric $D_{\mathcal{F}}(S,T)$ measures the difference between the distributions of the two domains $\mathcal{Z}^{(S)}$ and $\mathcal{Z}^{(T)}$. However, as addressed in Section \ref{sec:notation-setup}, if a domain $\mathcal{Z}^{(S)}$ differs from another domain $\mathcal{Z}^{(T)}$, there are three possibilities: the input-space distribution $\mathcal{D}^{(S)}$ differs from $\mathcal{D}^{(T)}$, or $g_*^{(S)}$ differs from $g_*^{(T)}$, or both of them occur. Therefore, it is necessary to consider two kinds of differences: the difference between the input-space distributions $\mathcal{D}^{(S)}$ and $\mathcal{D}^{(T)}$ and the difference between the labeling functions $g_*^{(S)}$ and $g_*^{(T)}$. Next, we will show that the integral probability metric $D_{\mathcal{F}}(S,T)$ can be bounded by using two separate quantities that can measure the difference between $\mathcal{D}^{(S)}$ and $\mathcal{D}^{(T)}$ and the difference between $g_*^{(S)}$ and $g_*^{(T)}$, respectively.

As shown in \eqref{eq:d-distance}, the quantity $\mathrm{disc}_{\ell}(\mathcal{D}^{(S)},\mathcal{D}^{(T)})$ actually measures the difference between the input-space distributions $\mathcal{D}^{(S)}$ and $\mathcal{D}^{(T)}$.
Moreover, we introduce another quantity to measure the difference between the labeling functions $g_*^{(S)}$ and $g_*^{(T)}$:
\begin{definition}\label{def:distance2}
Given a loss function $\ell$ and a function class $\mathcal{G}$, we define
\begin{equation}\label{eq:Q}
Q^{(T)}_{\mathcal{G}}(g_*^{(S)},g_*^{(T)})
:=\sup_{g_1\in\mathcal{G}}\Big|\mathrm{E}^{(T)}\ell\big(g_1({\bf x}^{(T)}),g_*^{(T)}({\bf x}^{(T)})\big)-\mathrm{E}^{(T)}\ell\big(g_1({\bf x}^{(T)}),g_*^{(S)}({\bf x}^{(T)})\big)\Big|.
\end{equation}
\end{definition}
Note that if the loss function $\ell$ and the function class $\mathcal{G}$ are both non-trivial ({\it i.e.}, $\mathcal{F}$ is non-trivial), the quantity $Q^{(T)}_{\mathcal{G}}(g_*^{(S)},g_*^{(T)})$ is a (semi)metric between the labeling functions $g_*^{(S)}$ and $g_*^{(T)}$. In fact, it is not hard to verify that $Q^{(T)}_{\mathcal{G}}(g_*^{(S)},g_*^{(T)})$ satisfies the triangle inequality and is equal to {\it zero} if $g_*^{(S)}$ and $g_*^{(T)}$ match.

By combining \eqref{eq:distance2}, \eqref{eq:d-distance} and \eqref{eq:Q}, we have
\begin{align}\label{eq:d1}
\mathrm{disc}_{\ell}(\mathcal{D}^{(S)},\mathcal{D}^{(T)})=&\sup_{g_1,g_2\in\mathcal{G}}\Big|\mathrm{E}^{(S)}\ell\big(g_1({\bf x}^{(S)}),g_2({\bf x}^{(S)})\big)-\mathrm{E}^{(T)}\ell\big(g_1({\bf x}^{(T)}),g_2({\bf x}^{(T)})\big)\Big|\nonumber\\
\geq&\sup_{g_1\in\mathcal{G}}\Big|\mathrm{E}^{(S)}\ell\big(g_1({\bf x}^{(S)}),g_*^{(S)}({\bf x}^{(S)})\big)-\mathrm{E}^{(T)}\ell\big(g_1({\bf x}^{(T)}),g_*^{(S)}({\bf x}^{(T)})\big)\Big|\nonumber\\
=&\sup_{g_1\in\mathcal{G}}\Big|\mathrm{E}^{(S)}\ell\big(g_1({\bf x}^{(S)}),g_*^{(S)}({\bf x}^{(S)})\big)-\mathrm{E}^{(T)}\ell\big(g_1({\bf x}^{(T)}),g_*^{(T)}({\bf x}^{(T)})\big)\nonumber\\
&+\mathrm{E}^{(T)}\ell\big(g_1({\bf x}^{(T)}),g_*^{(T)}({\bf x}^{(T)})\big)-\mathrm{E}^{(T)}\ell\big(g_1({\bf x}^{(T)}),g_*^{(S)}({\bf x}^{(T)})\big)\Big|\nonumber\\
\geq&\sup_{g_1\in\mathcal{G}}\Big|\mathrm{E}^{(S)}\ell\big(g_1({\bf x}^{(S)}),g_*^{(S)}({\bf x}^{(S)})\big)-\mathrm{E}^{(T)}\ell\big(g_1({\bf x}^{(T)}),g_*^{(T)}({\bf x}^{(T)})\big)\Big|\nonumber\\
&-\sup_{g_1\in\mathcal{G}}\Big|\mathrm{E}^{(T)}\ell\big(g_1({\bf x}^{(T)}),g_*^{(T)}({\bf x}^{(T)})\big)-\mathrm{E}^{(T)}\ell\big(g_1({\bf x}^{(T)}),g_*^{(S)}({\bf x}^{(T)})\big)\Big|\nonumber\\
=&D_{\mathcal{F}}(S,T)-Q^{(T)}_{\mathcal{G}}(g_*^{(S)},g_*^{(T)}),
\end{align}
and thus
\begin{equation}\label{eq:d2}
    D_{\mathcal{F}}(S,T)\leq
\mathrm{disc}_{\ell}(\mathcal{D}^{(S)},\mathcal{D}^{(T)})
+Q^{(T)}_{\mathcal{G}}(g_*^{(S)},g_*^{(T)}),
\end{equation}
which implies that the integral probability metric $D_{\mathcal{F}}(S,T)$ can be bounded by the summation of the {\it discrepancy distance} $\mathrm{disc}_{\ell}(\mathcal{D}^{(S)},\mathcal{D}^{(T)})$ and the quantity $Q^{(T)}_{\mathcal{G}}(g_*^{(S)},g_*^{(T)})$, which measure the difference between the input-space distributions $\mathcal{D}^{(S)}$ and $\mathcal{D}^{(T)}$ and the difference between the labeling functions $g_*^{(S)}$ and $g_*^{(T)}$, respectively.

\begin{remark}\label{rem:metric}
Note that there is a specific case in the situation of domain adaptation: $\mathcal{D}^{(S)}$ differs from $\mathcal{D}^{(T)}$ and meanwhile $g_*^{(S)}$ differs from $g_*^{(T)}$, while the distribution of the domain $\mathcal{Z}^{(S)}$ matches with that of the domain $\mathcal{Z}^{(T)}$. In this case, the integral probability metric $D_\mathcal{F}(S,T)$ equals to {\it zero}, but $\mathrm{disc}_{\ell}(\mathcal{D}^{(S)},\mathcal{D}^{(T)})$ or $Q^{(T)}_{\mathcal{G}}(g_*^{(S)},g_*^{(T)})$ neither equals to {\it zero}. Therefore, the integral probability metric $D_\mathcal{F}(S,T)$ is more suitable for this case than the discrepancy distance $\mathrm{disc}_{\ell}(\mathcal{D}^{(S)},\mathcal{D}^{(T)})$.
\end{remark}

\section{Complexity Measures of Function Classes}\label{sec:UEN}

Generally, the generalization bound of a certain learning process is achieved by incorporating some complexity measure of the function class, {\it e.g.}, the covering number, the VC dimension and the Rademacher complexity.
In this paper, we are mainly concerned with the uniform entropy number and the Rademacher complexity.

\subsection{Uniform Entropy Number}

The uniform entropy number is derived from the concept of the covering number and we refer to \citet{Mendelson03} for details. The covering number of a function class $\mathcal{F}$ is defined as follows:

\begin{definition}\label{def:CovNum}
Let $\mathcal{F}$ be a function class and $d$ be a metric on $\mathcal{F}$. For any $\xi>0$, the covering number of $\mathcal{F}$ at radius $\xi$
with respect to the metric $d$, denoted by $\mathcal{N}(\mathcal{F},\xi,d)$ is the minimum size of a cover of radius $\xi$.
\end{definition}
In some classical results of statistical learning theory, the covering number is applied by letting $d$ be the distribution-dependent metric. For example, as shown in Theorem 2.3 of \citet{Mendelson03}, one can set $d$ as the norm $\ell_1({\bf Z}_1^N)$ and then derive the generalization bound of the i.i.d. learning process by incorporating the expectation of the covering number, {\it i.e.,} $\mathrm{E}\mathcal{N}(\mathcal{F},\xi,\ell_1({\bf Z}_1^N))$.  However, in the situation of domain adaptation, we only know the information of the source domain, while the expectation $\mathrm{E}\mathcal{N}(\mathcal{F},\xi,\ell_1({\bf Z}_1^N))$ is dependent on the distributions of the source and the target domains because ${\bf z}=({\bf x},{\bf y})$. Therefore, the covering number is no longer suitable for our framework to obtain the generalization bounds for domain adaptation. In contrast, the uniform entropy number is distribution-free and thus we choose it as the complexity measure of function classes to derive the generalization bounds for domain adaptation.

Next, we will consider the uniform entropy number of $\mathcal{F}$ in the situations of two types of domain adaptation: (i) domain adaptation with multiple sources; (ii) domain adaptation combining source and target data, respectively.

\subsubsection{Domain Adaptation with Multiple Sources}

For clarity of presentation, we give a useful notation for the following discussion.
Let $\{{\bf Z}_{1}^{N_k}\}_{k=1}^K:=\{\{{\bf z}_{n}^{(k)}\}_{n=1}^{N_k}\}_{k=1}^K$ be the collection of sample sets drawn from multiple sources $\{\mathcal{Z}^{(S_k)}\}_{k=1}^K$, respectively. Denote $\{{\bf Z'}_{1}^{N_k}\}_{k=1}^K:=\{\{{\bf z'}_{n}^{(k)}\}_{n=1}^{N_k}\}_{k=1}^K$ as the collection of the ghost sample sets drawn from $\{\mathcal{Z}^{(S_k)}\}_{k=1}^K$ such that the ghost sample ${\bf z'}^{(k)}_{n}$ has the same distribution as ${\bf z}^{(k)}_{n}$ for any $1\leq k\leq K$ and any $1\leq n\leq N_k$. Denote ${\bf Z}_{1}^{2N_k}:=\{{\bf Z}_{1}^{N_k},{\bf Z'}_{1}^{N_k}\}$ for any $1\leq k\leq K$.  Moreover, given an $f\in\mathcal{F}$ and a ${\bf w}=(w_1,\cdots,w_K)\in[0,1]^K$ with $\sum_{k=1}^Kw_k=1$, we introduce a variant of the $\ell_1$ norm:
\begin{equation}\label{eq:L1_w}
  \|f\|_{\ell^{\bf w}_1(\{{\bf Z}_{1}^{2N_k}\}_{k=1}^K)}:=\sum_{k=1}^K\frac{w_k}{N_k}\sum_{n=1}^{N_k}|f({\bf z}_n^{(k)})|.
\end{equation}
It is noteworthy that the variant $\ell^{\bf w}_1$ of the $\ell_1$ norm is still a norm on the functional space, which can be directly verified by using the definition of norm, so we omit it here.

In the situation of domain adaptation with multiple sources, by setting the metric $d$ as $\ell^{\bf w}_1(\{{\bf Z}_{1}^{2N_k}\}_{k=1}^K)$, we then define the uniform entropy number of $\mathcal{F}$ with respect to the metric $\ell^{\bf w}_1(\{{\bf Z}_{1}^{2N_k}\}_{k=1}^K)$ as
\begin{equation}\label{eq:UEN1.M}
\ln\mathcal{N}_1^{\bf w}\big(\mathcal{F},\xi,2\sum_{k=1}^KN_k\big):=\sup_{\{{\bf Z}_{1}^{2N_k}\}_{k=1}^K}\ln\mathcal{N}\left(\mathcal{F},\xi,\ell^{\bf w}_1(\{{\bf Z}_{1}^{2N_k}\}_{k=1}^K)\right).
\end{equation}

\subsubsection{Domain Adaptation Combining Source and Target Data}

In the situation of domain adaptation combining source and target data, we have to introduce another variant of the $\ell_1$ norm on $\mathcal{F}$.
Let ${\bf Z}_{1}^{N_S}=\{{\bf z}_n^{(S)}\}_{n=1}^{N_S}$ and ${\bf \overline{Z}}_{1}^{N_T}=\{{\bf z}_n^{(T)}\}_{n=1}^{N_T}$ be two sets of samples drawn from the domains $\mathcal{Z}^{(S)}$ and $\mathcal{Z}^{(T)}$, respectively. Given an $f\in\mathcal{F}$, we define for any $\tau\in[0,1)$,
\begin{equation}\label{eq:L1_tau}
  \|f\|_{\ell^\tau_1({\bf Z}_{1}^{N_S},{\bf \overline{Z}}_{1}^{N_T})}:=\frac{\tau}{N_T}\sum_{n=1}^{N_T}|f({\bf z}_n^{(T)})|+\frac{1-\tau}{N_S}\sum_{n=1}^{N_S}|f({\bf z}_n^{(S)})|.
\end{equation}
Note that the variant $\ell_1^\tau$ ($\tau\in[0,1)$) of the norm $\ell_1$ is still a norm on the functional space, which can be easily verified by using the definition of norm, so we omit it here.

Moreover, let ${\bf Z'}_{1}^{N_S}$ and ${\bf \overline{Z}'}_{1}^{N_T}$ be the ghost sample sets of ${\bf Z}_{1}^{N_S}$ and ${\bf \overline{Z}}_{1}^{N_T}$, respectively. Denote ${\bf Z}_{1}^{2N_{S}}:=\{{\bf Z}_{1}^{N_S},{\bf Z'}_{1}^{N_S}\}$ and ${\bf \overline{Z}}_{1}^{2N_T}:=\{{\bf \overline{Z}}_{1}^{N_T},{\bf \overline{Z}'}_{1}^{N_T}\}$, respectively. Then, the uniform entropy number of $\mathcal{F}$ with respect to the metric $\ell_1^\tau({\bf Z})$ is defined as
\begin{equation}\label{eq:UEN2.C}
\ln\mathcal{N}_1^{\tau}(\mathcal{F},\xi,2(N_S+N_T)):=\sup_{{\bf Z}}\ln\mathcal{N}\left(\mathcal{F},\xi,\ell_1^{\tau}({\bf Z})\right),
\end{equation}
where ${\bf Z}:=\{{\bf Z}_{1}^{2N_S},{\bf \overline{Z}}_{1}^{2N_T}\}$.

\subsection{Rademacher Complexity}

The Rademacher complexity is one of the most frequently used complexity measures of function classes and we refer to \citet{Vaart96,Mendelson03} for details.

\begin{definition}\label{def:Rade}
Let $\mathcal{F}$ be a function class and $\{{\bf z}_n\}_{n=1}^N$ be a sample set drawn from $\mathcal{Z}$. Denote $\{\sigma_n\}_{n=1}^N$ be a set of random variables independently taking either value from $\{-1,1\}$ with equal probability. Rademacher complexity of $\mathcal{F}$ is defined as
\begin{equation}\label{eq:ExRade}
\mathcal{R}(\mathcal{F}):=\mathrm{E}\sup_{f\in\mathcal{F}}
\left\{\frac{1}{N}\big|\sum_{n=1}^N\sigma_nf({\bf z}_n)\big|\right\}
\end{equation}
with its empirical version
\begin{equation}\label{eq:EmRade}
\mathcal{R}_N(\mathcal{F}):=\mathrm{E}_{\sigma}\sup_{f\in\mathcal{F}}
\left\{\frac{1}{N}\big|\sum_{n=1}^N\sigma_nf({\bf z}_n)\big|\right\},
\end{equation}
where $\mathrm{E}$ stands for the expectation taken with respect to all random variables $\{{\bf z}_n\}_{n=1}^N$ and $\{\sigma_n\}_{n=1}^N$, and $\mathrm{E}_{\sigma}$ stands for the expectation only taken with respect to the random variables $\{\sigma_n\}_{n=1}^N$.
\end{definition}


\section{Learning Processes of Domain Adaptation with Multiple Sources}\label{sec:MS}

In this section, we present two generalization bounds of the learning process for domain adaptation with multiple sources. They are based on the uniform entropy number and the Rademacher complexity, respectively. By using the derived bounds based on the uniform entropy number, we then analyze the asymptotic convergence and the rate of convergence of the learning process. The numerical experiment supports our theoretical analysis as well.

\subsection{Generalization Bounds}

Based on the uniform entropy number defined in \eqref{eq:UEN1.M}, a generalization bound for domain adaptation with multiple sources is presented in the following theorem.

\begin{theorem}\label{thm:crate.M}
Assume that $\mathcal{F}$ is a function class consisting of bounded functions with the range $[a,b]$. Let ${\bf w}=(w_1,\cdots,w_K)\in[0,1]^K$ with $\sum_{k=1}^Kw_k=1$.
Then, given an arbitrary $\xi>D^{({\bf w})}_{\mathcal{F}}(S,T)$,
 we have  for any $\big(\prod_{k=1}^KN_k\big)\geq\frac{8\left(b-a\right)^2}{(\xi')^2}$ and any $\epsilon>0$, with probability at least $1-\epsilon$,
\begin{align}\label{eq:crate1.M}
\sup_{f\in\mathcal{F}}
\big|\mathrm{E}^{(S)}_{\bf w}f-\mathrm{E}^{(T)}f\big|
\leq D^{({\bf w})}_{\mathcal{F}}(S,T)+ \left(\frac{\left(\ln\mathcal{N}_1^{\bf w}\big(\mathcal{F},\xi'/8,2\sum_{k=1}^KN_k\big)-\ln(\epsilon/8)\right)}
{\frac{\big(\prod_{k=1}^KN_k\big)}{32(b-a)^2\big(\sum_{k=1}^{K}
w_k^2(\prod_{i\not=k}N_i)\big)}}\right)^{\frac{1}{2}},
\end{align}
where $\xi'=\xi-D^{({\bf w})}_{\mathcal{F}}(S,T)$ and
\begin{equation}\label{eq:Dist.M}
D^{({\bf w})}_{\mathcal{F}}(S,T):=\sum_{k=1}^Kw_kD_{\mathcal{F}}(S_k,T).
\end{equation}
\end{theorem}

In the above theorem, we show that the generalization bound $\sup_{f\in\mathcal{F}}|\mathrm{E}^{(T)}f-\mathrm{E}^{(S)}_{\bf w}f|$ can be bounded by the right-hand side of \eqref{eq:crate1.M}.
Compared to the classical result under the assumption of same distribution \citep[see][Theorem 2.3 and Definition 2.5]{Mendelson03}: with probability at least $1-\epsilon$,
\begin{align}\label{eq:crate2.1}
    \sup_{f\in\mathcal{F}}
\big|\mathrm{E}_Nf-\mathrm{E}f\big|
\leq O\left(\left(\frac{\ln\mathcal{N}_1\big(\mathcal{F},\xi,N\big)-\ln(\epsilon/8)}{N}\right)^{\frac{1}{2}}\right)
\end{align}
with $\mathrm{E}_Nf$ being the empirical risk with respect to the  sample set ${\bf Z}_1^N$, there is a discrepancy quantity $D^{({\bf w})}_{\mathcal{F}}(S,T)$ that is determined by the two factors: the choice of ${\bf w}$ and the integral probability metrics $D_{\mathcal{F}}(S_k,T)$ ($1\leq k\leq K$). The two results will coincide if any source domain and the target domain match, {\it i.e.}, $D_{\mathcal{F}}(S_k,T)=0$ holds for any $1\leq k\leq K$.

In order to prove this result, we develop the specific Hoeffding-type deviation inequality and the symmetrization inequality for domain adaptation with multiple sources, respectively. The detailed proof is arranged in Appendix \ref{app:Proof1}. Subsequently, we give another generalization bound based on the Rademacher complexity:

\begin{theorem}\label{thm:RB.Rade}
Assume that $\mathcal{F}$ is a function class consisting of bounded functions with the range $[a,b]$. Let ${\bf w}=(w_1,\cdots,w_K)\in[0,1]^K$ with $\sum_{k=1}^Kw_k=1$.
Then, we have with probability at least $1-\epsilon$,
\begin{align}\label{eq:RB.Rade}
\sup_{f\in\mathcal{F}}\big|\mathrm{E}_{{\bf w}}^{(S)}f-\mathrm{E}^{(T)}f\big|\leq D^{({\bf w})}_{\mathcal{F}}(S,T)+2\sum_{k=1}^Kw_k\mathcal{R}^{(k)}(\mathcal{F})+
\sqrt{\sum_{k=1}^K\frac{(b-a)^2w_k^2\ln(1/\epsilon)}{2N_k}},
\end{align}
where $D^{({\bf w})}_{\mathcal{F}}(S,T)$ is defined in \eqref{eq:Dist.M} and $\mathcal{R}^{(k)}(\mathcal{F})$ ($1\leq k\leq K$) are the Rademacher complexities on the source domains $\mathcal{Z}^{(S_k)}$, respectively.
\end{theorem}
Similarly, the derived bound \eqref{eq:RB.Rade} coincides with the related classical result under the assumption of same distribution \citep[see][Theorem 5]{Bousquet04}, when any source domain of $\{\mathcal{Z}^{(S_k)}\}_{k=1}^K$ and the target domain $\mathcal{Z}^{(T)}$ match, {\it i.e.}, $D^{({\bf w})}_{\mathcal{F}}(S,T)=D_{\mathcal{F}}(S_k,T)=0$ holds for any $1\leq k\leq K$.
The proof of this theorem is processed by introducing a generalized version of McDiarmid's inequality which allows independent random variables to take values from different domains (see Appendix \ref{app:Proof3}).

Subsequently, based on the derived bound \eqref{eq:crate1.M}, we can analyze the asymptotic behavior of the learning process for domain adaptation with multiple sources.


\subsection{Asymptotic Convergence}

In statistical learning theory, it is well-known that the complexity of function class is one of main factors to the asymptotic convergence of the learning process under the assumption of same distribution  \citep{Vapnik99,Vaart96,Mendelson03}.

From Theorem \ref{thm:crate.M}, we can directly arrive at the following result showing that the asymptotic convergence of the learning process for domain adaptation with multiple sources is affected by the three aspects: the choice of ${\bf w}$, the discrepancy quantity $D^{({\bf w})}_{\mathcal{F}}(S,T)$ and the uniform entropy number $\ln\mathcal{N}^{\bf w}_1\big(\mathcal{F},\xi/8,2\sum_{k=1}^KN_k\big)$.

\begin{theorem}\label{thm:converge.M}
Assume that $\mathcal{F}$ is a function class consisting of the bounded functions with the range $[a,b]$. Let ${\bf w}=(w_1,\cdots,w_K)\in[0,1]^K$ with $\sum_{k=1}^Kw_k=1$. If the following condition holds:
\begin{equation}\label{eq:cong1.M}
  \lim_{N_1,\cdots,N_K\rightarrow +\infty}\frac{\ln\mathcal{N}^{\bf w}_1\big(\mathcal{F},\xi/8,2\sum_{k=1}^KN_k\big)}
{\frac{\big(\prod_{k=1}^KN_k\big)}{32(b-a)^2\big(\sum_{k=1}^{K}w_k^2(\prod_{i\not=k}N_i)\big)}}<+\infty,
\end{equation}
then we have for any $\xi>D^{({\bf w})}_{\mathcal{F}}(S,T)$,
\begin{equation}\label{eq:converge.M}
    \lim_{N_1,\cdots,N_K\rightarrow+\infty}\mathrm{Pr}
    \Big\{\sup_{f\in\mathcal{F}}\big|\mathrm{E}^{(T)}f-\mathrm{E}_{\bf w}^{(S)}f\big|>\xi\Big\}=0.
\end{equation}
\end{theorem}

As shown in Theorem \ref{thm:converge.M}, if the choice of ${\bf w}\in[0,1]^K$ and the uniform entropy number $\ln\mathcal{N}^{\bf w}_1\big(\mathcal{F},\xi'/8,2\sum_{k=1}^KN_k\big)$ satisfy the condition \eqref{eq:cong1.M} with $\sum_{k=1}^Kw_k=1$,
the probability of the event that $\sup_{f\in\mathcal{F}}\big|\mathrm{E}^{(T)}f-\mathrm{E}_{\bf w}^{(S)}f\big|>\xi$ will converge to {\it zero} for any $\xi>D^{({\bf w})}_{\mathcal{F}}(S,T)$, when the sample numbers $N_1,\cdots,N_K$ of multiple sources go to {\it infinity}, respectively. This is partially in accordance with the classical result of the asymptotic convergence
of the learning process under the assumption of same distribution ({\it cf.} Theorem 2.3 and Definition 2.5 of [22]): the probability of the event that $\sup_{f\in\mathcal{F}}\big|\mathrm{E}f-\mathrm{E}_Nf\big|>\xi$ will converge to {\it zero} for any $\xi>0$, if the uniform entropy number $\ln\mathcal{N}_1\left(\mathcal{F},\xi,N\right)$ satisfies the following:
\begin{equation}\label{eq:cong2}
  \lim_{N\rightarrow +\infty}\frac{\ln\mathcal{N}_1\left(\mathcal{F},\xi,N\right)}{N}<+\infty.
\end{equation}

Note that in the learning process of domain adaptation with multiple sources, the uniform convergence of the empirical risk on the source domains to the expected risk on the target domain may not hold, because the limit \eqref{eq:converge.M} does not hold for any $\xi>0$ but for any $\xi>D^{({\bf w})}_{\mathcal{F}}(S,T)$. By contrast, the limit \eqref{eq:converge.M} holds for all $\xi>0$ in the learning process under the assumption of same distribution, if the condition \eqref{eq:cong2} is satisfied. Again, these two results coincide when any source domain and the target domain match, {\it i.e.}, $D^{({\bf w})}_{\mathcal{F}}(S,T)=D_{\mathcal{F}}(S_k,T)=0$ holds for any $1\leq k\leq K$.

 Next, we study the rate of convergence of the learning process for domain adaptation with multiple sources.

\subsection{Rate of Convergence}
Recalling \eqref{eq:crate1.M}, we can find that the rate of convergence is affected by the choice of ${\bf w}$.
According to the Cauchy-Schwarz inequality, setting $w_k=N_k/\sum_{k=1}^KN_k$ ($1\leq k\leq K$),
we have
\begin{equation}\label{eq:rate.max}
 \max\left\{\frac{\big(\prod_{k=1}^KN_k\big)}{32(b-a)^2
\big(\sum_{k=1}^{K}w_k^2(\prod_{i\not=k}N_i)\big)}\right\}
=\frac{N_1+N_2+\cdots+N_K}{32(b-a)^2},
\end{equation}
which minimizes the second term of the right-hand side of \eqref{eq:crate1.M}.
Thus, by \eqref{eq:crate1.M}, \eqref{eq:crate2.1} and \eqref{eq:rate.max}, we find that the fastest rate of convergence of the learning process is up to $O(1/\sqrt{N})$ which is the same as the classical result \eqref{eq:crate2.1} of the learning process under the assumption of same distribution if the discrepancy $D^{({\bf w})}_{\mathcal{F}}(S,T)$ is ignored.

In addition, the bound \eqref{eq:RB.Rade} based on the Rademacher complexity also implies that the rate of convergence of the learning process is affected by the choice of ${\bf w}$. Again, according to Cauchy-Schwarz inequality, setting $w_k=\frac{N_k}{\sum_{k=1}^KN_k}$ ($1\leq k\leq K$) leads to the fastest rate of convergence:
\begin{equation*}
 \sqrt{\frac{(b-a)^2\ln(1/\epsilon)}{2\sum_{k=1}^KN_k}}=O(1/\sqrt{N}),
\end{equation*}
which is in accordance with the aforementioned analysis. The following numerical experiments support our theoretical findings (see Fig. \ref{fig:fig1}).

\subsection{Numerical Experiment}

We have performed the numerical experiments to verify the theoretic analysis of the asymptotic convergence of the learning processes for domain adaptation with multiple sources. Without loss of generality, we only consider the case of $K=2$, {\it i.e.,} there are two source domains and one target domain. The experiment data are generated in the following way:

 For the target domain $\mathcal{Z}^{(T)}=\mathcal{X}^{(T)}\times\mathcal{Y}^{(T)}\subset\mathbb{R}^{100}\times\mathbb{R}$, we consider $\mathcal{X}^{(T)}$ as a Gaussian distribution $N(0,1)$ and draw $\{{\bf x}^{(T)}_n\}_{n=1}^{N_T}$ ($N_T=4000$) from $\mathcal{X}^{(T)}$ randomly and independently.
  Let $\beta\in\mathbb{R}^{100}$ be a random vector of a Gaussian distribution $N(1,5)$, and let the random vector $R\in\mathbb{R}^{100}$ be a noise term with $R\sim N(0,0.5)$.
 For any $1\leq n\leq N_T$, we randomly draw $\beta$ and $R$ from $N(1,5)$ and $N(0,0.01)$ respectively, and then generate $y^{(T)}_n\in\mathcal{Y}$ as follows:
\begin{equation}\label{eq:MS.T}
y^{(T)}_n=\langle{\bf x}^{(T)}_n, \beta\rangle+R.
\end{equation}
The derived $\{({\bf x}_n^{(T)}, y_n^{(T)})\}_{n=1}^{N_T}$ ($N_T=4000$) are the samples of the target domain $\mathcal{Z}^{(T)}$ and will be used as the test data.

In the similar way, we generate the sample set $\{({\bf x}_n^{(1)}, y_n^{(1)})\}_{n=1}^{N_1}$ ($N_1=2000$) of the source domain $\mathcal{Z}^{(S_1)}=\mathcal{X}^{(1)}\times\mathcal{Y}^{(1)}\subset\mathbb{R}^{100}\times\mathbb{R}$: for any $1\leq n\leq N_1$,
\begin{equation}\label{eq:MS.1}
y^{(1)}_n=\langle{\bf x}^{(1)}_n, \beta\rangle+R,
\end{equation}
where ${\bf x}^{(1)}_n\sim N(0.5,1)$, $\beta \sim N(1,5)$ and $R\sim N(0,0.5)$.

For the source domain $\mathcal{Z}^{(S_2)}=\mathcal{X}^{(2)}\times\mathcal{Y}^{(2)}\subset\mathbb{R}^{100}\times\mathbb{R}$, the samples  $\{({\bf x}_n^{(2)}, y_n^{(2)})\}_{n=1}^{N_2}$ ($N_2=2000$) are derived in the following way: for any $1\leq n\leq N_2$,
\begin{equation}\label{eq:MS.2}
y^{(2)}_n=\langle{\bf x}^{(2)}_n, \beta\rangle+R,
\end{equation}
where ${\bf x}^{(2)}_n\sim N(2,5)$, $\beta \sim N(1,5)$ and $R\sim N(0,0.5)$.

\begin{figure}[htbp]
\begin{center}
\includegraphics[height=8cm]{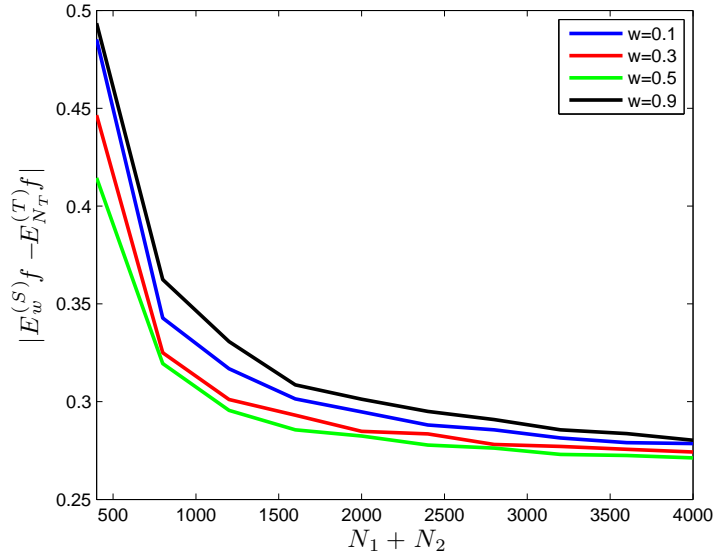}
\caption{Domain Adaptation with Multiple Sources}
\label{fig:fig1}
\end{center}
\end{figure}

In this experiment, we use the method of Least Square Regression\footnote{SLEP Package: http://www.public.asu.edu/$\sim$jye02/Software/SLEP/index.htm} to minimize the empirical risk
\begin{equation}\label{eq:experiment1}
\mathrm{E}^{(S)}_w(\ell\circ g)=\frac{w}{N_1}\sum_{n=1}^{N_1}\ell(g({\bf x}^{(1)}_n),y^{(1)}_n)+\frac{(1-w)}{N_2}\sum_{n=1}^{N_2}\ell(g({\bf x}^{(2)}_n),y^{(2)}_n)
 \end{equation}
for different combination coefficients $w\in\{0.1,0.3,0.5,0.9\}$ and then compute the discrepancy $|E^{(S)}_wf-E^{(T)}_{N_T}f|$ for each $N_1+N_2$. The initial $N_1$ and $N_2$ both equal to $200$. Each test is repeated $30$ times and the final result is the average of the $30$ results.
After each test, we increment $N_1$ and $N_2$ by $200$ until $N_1=N_2=2000$. The experiment results are shown in Fig. \ref{fig:fig1}.

From Fig. \ref{fig:fig1}, we can find that for any choice of $w$, the curve of $|E^{(S)}_wf-E^{(T)}_{N_T}f|$ is decreasing when $N_1+N_2$ increases, which is in accordance with the results presented in Theorems \ref{thm:crate.M} \& \ref{thm:converge.M}. Moreover, when $w=0.5$, the discrepancy $|E^{(S)}_wf-E^{(T)}_{N_T}f|$ has the fastest rate of convergence, and the rate becomes slower as $w$ is further away from $0.5$. In this experiment, we set $N_1=N_2$ that implies that $N_2/(N_1+N_2)=0.5$. Recalling \eqref{eq:crate1.M}, we have shown that $w=N_2/(N_1+N_2)$ will provide the fastest rate of convergence and this proposition is supported by the experiment results shown in Fig. \ref{fig:fig1}.


\section{Learning Process of Domain Adaptation Combining Source and Target Data}\label{sec:CSTD}

In this section, we present two generalization bounds of the learning process for domain adaptation combining source and target data, which are based on the uniform entropy number and the Rademacher complexity, respectively. We then analyze the asymptotic convergence and the rate of convergence of the learning process in addition to the numerical experiments supporting our theoretical analysis.

\subsection{Generalization Bounds}

The following theorem provides a generalization bound based on the uniform entropy number with respect to the metric $\ell_1^{\tau}$ defined in \eqref{eq:UEN2.C}. Similar to the situation of domain adaptation with multiple sources, the proof of this theorem is achieved by using a specific Hoeffding-type deviation inequality and a symmetrization inequality for domain adaptation combining source and target data (see Appendix \ref{app:Proof2}).

\begin{theorem}\label{thm:crate.C}
Assume that $\mathcal{F}$ is a function class consisting of the bounded functions with the range $[a,b]$. Let ${\bf Z}_{1}^{N_S}=\{{\bf z}_n^{(S)}\}_{n=1}^{N_S}$ and ${\bf \overline{Z}}_{1}^{N_T}=\{{\bf z}_n^{(T)}\}_{n=1}^{N_T}$ be two sets of i.i.d. samples drawn from domains $\mathcal{Z}^{(S)}$ and $\mathcal{Z}^{(T)}$, respectively.
Then, for any $\tau\in[0,1)$ and given an arbitrary $\xi>(1-\tau)D_{\mathcal{F}}(S,T)$,
we have for any $N_SN_T\geq\frac{8(b-a)^2}{(\xi')^2}$, with probability at least $1-\epsilon$,
\begin{align}\label{eq:crate1.C}
   \sup_{f\in\mathcal{F}}
\big|\mathrm{E}_\tau f-\mathrm{E}^{(T)}f\big|
\leq(1-\tau)D_{\mathcal{F}}(S,T)
+ \left(\frac{\ln\mathcal{N}_1^{\tau}(\mathcal{F},\xi'/8,2(N_S+N_T))-\ln(\epsilon/8)}
{\frac{N_SN_T}{32(b-a)^2\left((1-\tau)^2N_T+\tau^2N_S\right)}}\right)^{\frac{1}{2}},
\end{align}
where $D_{\mathcal{F}}(S,T)$ is defined in \eqref{eq:distance} and $\xi':=\xi-(1-\tau)D_{\mathcal{F}}(S,T)$.
\end{theorem}

Compared to the classical result \eqref{eq:crate2.1} under the assumption of same distribution, the derived bound \eqref{eq:crate1.C} contains a term of discrepancy quantity $(1-\tau)D_{\mathcal{F}}(S,T)$ that is determined by two factors:
the combination coefficient $\tau$ and the quantity $D_{\mathcal{F}}(S,T)$.
The two results coincide when the source domain $\mathcal{Z}^{(S)}$ and the
target domain $\mathcal{Z}^{(T)}$ match, {\it i.e.,} $D_{\mathcal{F}}(S,T)=0$.

Based on the Rademacher complexity, we then get another generalization bound of the learning process for domain adaptation combining source and target data.
Its proof is postponed in Appendix \ref{app:Proof3}.

\begin{theorem}\label{thm:RB.Rade.C}
Assume that $\mathcal{F}$ is a function class consisting of the bounded functions with the range $[a,b]$. Let ${\bf Z}_{1}^{N_S}=\{{\bf z}_n^{(S)}\}_{n=1}^{N_S}$ and $\overline{{\bf Z}}_{1}^{N_T}=\{{\bf z}_n^{(T)}\}_{n=1}^{N_T}$ be two sets of i.i.d. samples drawn from the domains $\mathcal{Z}^{(S)}$ and $\mathcal{Z}^{(T)}$, respectively.
Then, given $\tau\in[0,1)$ and for any $\epsilon>0$,
we have with probability at least $1-\epsilon$,
\begin{align}\label{eq:RB.Rade.C}
\sup_{f\in\mathcal{F}}\big|\mathrm{E}_{\tau}f-\mathrm{E}^{(T)}f\big|\leq& (1-\tau)D_{\mathcal{F}}(S,T)+2(1-\tau)\mathcal{R}^{(S)}(\mathcal{F})\nonumber\\
&+2\tau\mathcal{R}^{(T)}_{N_T}(\mathcal{F})+
3
\tau\sqrt{\frac{(b-a)\ln(4/\epsilon)}{2N_T}}\nonumber\\
&+(1-\tau)
\sqrt{\frac{(b-a)^2\ln(2/\epsilon)}{2}\left(\frac{\tau^2}{N_T}+\frac{(1-\tau)^2}{N_S}\right)},
\end{align}
where $D_{\mathcal{F}}(S,T)$ is defined in \eqref{eq:distance}.
\end{theorem}
Note that in the derived bound \eqref{eq:RB.Rade.C}, we adopt an empirical Rademacher complexity $\mathcal{R}^{(T)}_{N_T}(\mathcal{F})$ that is based on the data drawn from the target domain $\mathcal{Z}^{(T)}$, because the distribution of $\mathcal{Z}^{(T)}$ is unknown in the situation of domain adaptation.
Similar to the aforementioned discussion, the generalization bound \eqref{eq:RB.Rade.C} coincides with the result under the assumption of same distribution \citep[see][Theorem 5]{Bousquet04}, when the source domain of $\mathcal{Z}^{(S)}$ and the target domain $\mathcal{Z}^{(T)}$ match, {\it i.e.}, $D_{\mathcal{F}}(S,T)=0$.

The two results \eqref{eq:crate1.C} and \eqref{eq:RB.Rade.C} exhibit a tradeoff between the sample numbers $N_S$ and $N_T$, which is associated with the choice of $\tau$. Although the tradeoff has been mentioned in some previous works \citep[see][]{Blitzer08,Ben-David10}, the following will show a rigorous theoretical analysis of the tradeoff.


\subsection{Asymptotic Convergence}

Following Theorem \ref{thm:crate.C}, we can directly obtain the concerning result pointing out that the asymptotic convergence of the learning process for domain adaptation combining source and target data is affected by three factors: the uniform entropy number $\ln\mathcal{N}_{1}^{\tau}(\mathcal{F},\xi/8,2(N_S+N_T))$, the integral probability metric $D_{\mathcal{F}}(S,T)$ and the choice of $\tau\in[0,1)$.

\begin{theorem}\label{thm:converge.C}
Assume that $\mathcal{F}$ is a function class consisting of bounded functions with the range $[a,b]$. Given a $\tau\in[0,1)$, if the following condition holds:
\begin{equation}\label{eq:cong1.C}
  \lim_{N_S\rightarrow +\infty}\frac{\ln\mathcal{N}_1^{\tau}(\mathcal{F},\xi'/8,2(N_S+N_T))}{\frac{N_SN_T}
  {\left((1-\tau)^2N_T+\tau^2N_S\right)}}<+\infty
\end{equation}
with $\xi':=\xi-(1-\tau)D_{\mathcal{F}}(S,T)$,
then we have for any $\xi>(1-\tau)D_{\mathcal{F}}(S,T)$,
\begin{equation}\label{eq:converge.C}
    \lim_{N_S\rightarrow+\infty}\mathrm{Pr}\left\{\sup_{f\in\mathcal{F}}
    \big|\mathrm{E}^{(T)}f-\mathrm{E}_\tau f\big|>\xi\right\}=0.
\end{equation}
\end{theorem}

As shown in Theorem \ref{thm:converge.C}, if the choice of $\tau\in[0,1)$ and the uniform entropy number $\ln\mathcal{N}_1^\tau(\mathcal{F},\xi'/8,2(N_S+N_T))$ satisfy the condition \eqref{eq:cong1.C},
the probability of the event $\sup_{f\in\mathcal{F}}\big|\mathrm{E}^{(T)}f-\mathrm{E}_\tau f\big|>\xi$ will converge to {\it zero} for any $\xi>(1-\tau)D_{\mathcal{F}}(S,T)$, when $N_S$ goes to {\it infinity}. This is partially in accordance with the classical result under the assumption of same distributions derived from the combination of Theorem 2.3 and Definition 2.5 of \citet{Mendelson03}.

Note that in the learning process for domain adaptation combining source and target data, the uniform convergence of the empirical risk $\mathrm{E}_\tau f$ to the expected risk $\mathrm{E}^{(T)}f$ may not hold, because the limit \eqref{eq:converge.C} does not hold for any $\xi>0$ but for any $\xi>(1-\tau)D_{\mathcal{F}}(S,T)$. By contrast, the limit \eqref{eq:converge.C} holds for all $\xi>0$ in the learning process under the assumption of same distribution, if the condition \eqref{eq:cong2} is satisfied. The two results coincide when the source domain $\mathcal{Z}^{(S)}$ and the
target domain $\mathcal{Z}^{(T)}$ match, {\it i.e.,} $D_{\mathcal{F}}(S,T)=0$.

\subsection{Rate of Convergence}

We consider the choice of $\tau$ that is an essential factor to the rate of convergence for the learning process and is associated with the tradeoff between the sample numbers $N_S$ and $N_T$.
Recalling \eqref{eq:crate1.C}, if we fix the value of $\ln\mathcal{N}_1^\tau(\mathcal{F},\xi'/8,2(N_S+N_T))$, setting $\tau=\frac{N_T}{N_T+N_S}$ minimizes the second term of the right-hand side of \eqref{eq:crate1.C} and then we arrive at
  \begin{align}\label{eq:crate1.C2}
    \sup_{f\in\mathcal{F}}
\big|\mathrm{E}_\tau f-\mathrm{E}^{(T)}f\big|
\leq\frac{N_SD_{\mathcal{F}}(S,T)}{N_S+N_T}+ \left(\frac{\left(\ln\mathcal{N}_1^\tau(\mathcal{F},\xi'/8,2(N_S+N_T))-\ln(\epsilon/8)\right)}
{\frac{N_S+N_T}{32(b-a)^2}}\right)^{\frac{1}{2}},
\end{align}
which implies that setting $\tau=\frac{N_T}{N_T+N_S}$ can result in the fastest rate of convergence, while it can also cause the relatively larger discrepancy between the empirical risk $\mathrm{E}_\tau f$ and the expected risk $\mathrm{E}^{(T)}f$, because the situation of domain adaptation is set up in the condition that $N_T\ll N_S$, which implies that $\frac{N_S}{N_S+N_T}\approx1$. Moreover, this choice of $\tau$ associated with a trade off between sample numbers $N_S$ and $N_T$ is also suitable to the Rademacher-complexity-based bound \eqref{eq:RB.Rade.C}.
It is noteworthy that the value $\tau=\frac{N_T}{N_T+N_S}$ has been mentioned in the section of ``Experimental Results" in \citet{Blitzer08}. Here, we show a rigorous theoretical analysis of this value and the following numerical experiment also supports this finding (see Fig. \ref{fig:fig2}).

\subsection{Numerical Experiments}

In the situation of domain adaptation combining source and target data, the samples $\{({\bf x}_n^{(T)}, y_n^{(T)})\}_{n=1}^{N_T}$ ($N_T=4000$) of the target domain $\mathcal{Z}^{(T)}$ are generated in the aforementioned way (see \eqref{eq:MS.T}). We randomly pick $N'_T=100$ samples from them to form the objective function and the rest $N''_T=3900$ are used to test.

In the similar way, the samples $\{({\bf x}_n^{(S)}, y_n^{(S)})\}_{n=1}^{N_S}$ ($N_S=4000$) of the source domain $\mathcal{Z}^{(S)}$ are generated as follows: for any $1\leq n\leq N_S$,
\begin{equation}\label{eq:CST}
y^{(S)}_n=\langle{\bf x}^{(S)}_n, \beta\rangle+R,
\end{equation}
where ${\bf x}^{(S)}_n\sim N(1,2)$, $\beta \sim N(1,5)$ and $R\sim N(0,0.5)$.

\begin{figure}[htbp]
\begin{center}
\includegraphics[height=8cm]{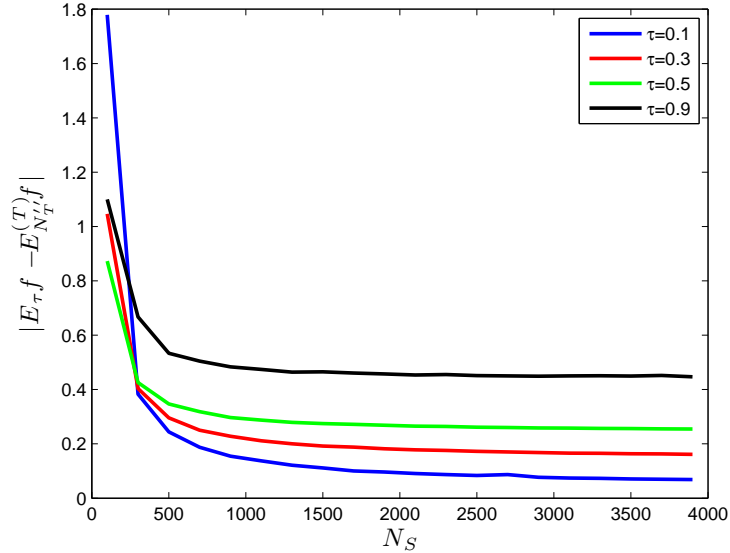}
\caption{Domain Adaptation Combining Source and Target Data}
\label{fig:fig2}
\end{center}
\end{figure}

We also use the method of Least Square Regression to minimize the empirical risk
\begin{equation*}
\mathrm{E}_\tau(\ell\circ g)=\frac{1-\tau}{N_S}\sum_{n=1}^{N_S}\ell(g({\bf x}^{(1)}_n),y^{(1)}_n)+\frac{\tau}{N'_T}\sum_{n=1}^{N'_T}\ell(g({\bf x}^{(T)}_n),y^{(T)}_n)
 \end{equation*}
for different combination coefficients $\tau\in\{0.1,0.3,0.5,0.9\}$ and then compute the discrepancy $|E_\tau f-E^{(T)}_{N''_T}f|$ for each $N_S$. Since it has to be satisfied that $N_S\gg N'_T$, the initial $N_S$ is set to be $200$. Each test is repeated $100$ times and the final result is the average of the $100$ results.
After each test, we increment $N_S$ by $200$ until $N_S=4000$. The experiment results are shown in Fig. \ref{fig:fig2}.

Figure \eqref{fig:fig2} illustrates that for any choice of $\tau\in\{0.1,0.3,0.5,0.9\}$, the curve of $|E_\tau f-E^{(T)}_{N''_T}|$ is decreasing as $N_S$ increases. This is in accordance with our results of the asymptotic convergence of the learning process for domain adaptation with multiple sources  (see Theorems \ref{thm:crate.C} and \ref{thm:converge.C}). Furthermore, Fig. \ref{fig:fig2} also shows that when $\tau \approx N'_T/(N_S+N'_T)$, the discrepancy $|E^{(S)}_\tau f-E^{(T)}_{N''_T}f|$ has the fastest rate of convergence, and
the rate becomes slower as $\tau$ is further away from $N'_T/(N_S+N'_T)$. Thus, this is in accordance with the theoretical analysis of the asymptotic convergence presented above.


\section{Prior Works}\label{sec:compare}

There have been some previous works on the theoretical analysis of domain adaptation with multiple sources \citep[see][]{Ben-David10,Crammer06,Crammer08,Mansour08,Mansour09} and domain adaptation combining source and target data \citep[see][]{Blitzer08,Ben-David10}.

In \citet{Crammer06,Crammer08}, the function class and the loss function are assumed to satisfy the conditions of ``$\alpha$-triangle inequality" and ``uniform convergence bound". Moreover, one has to get some prior information about the disparity between any source domain and the target domain. Under these conditions, some generalization bounds were obtained by using the classical techniques developed under the assumption of same distribution.

\citet{Mansour08} proposed another framework to study the problem of domain adaptation with multiple sources. In this framework, one has to know some prior knowledge including the exact distributions of the source domains and the hypothesis function with a small loss on each source domain. Furthermore, the target domain and the hypothesis function on the target domain were deemed as the mixture of the source domains and the mixture of the
hypothesis functions on the source domains, respectively. Then, by introducing the R\'enyi divergence, \citet{Mansour09} extended their previous work \citep{Mansour08} to a more general setting, where the distribution of the target domain can be arbitrary and one only needs to know an approximation of the exact distribution of each source domain.
\citet{Ben-David10} also discussed the situation of domain adaptation with the mixture of source domains.

In \cite{Ben-David10,Blitzer08}, domain adaptation combining source and target data was originally proposed and meanwhile a theoretical framework was presented to analyze its properties for the classification tasks by introducing the $\mathcal{H}$-divergence. Under the condition of ``$\lambda$-close", the authors applied the classical techniques developed under the assumption of same distribution to achieve the generalization bounds based on the VC dimension.

\citet{Mansour09b} introduced the {\it discrepancy distance} $\mathrm{disc}_{\ell}(\mathcal{D}^{(S)},\mathcal{D}^{(T)})$ to capture the difference between domains and this quantity can be used in both classification and regression tasks. By applying the classical results of statistical learning theory, the authors obtained the generalization bounds based on the Rademacher complexity.



\section{Conclusion}

In this paper, we propose a new framework to obtain generalization bounds of the learning process for two representative types of domain adaptation: domain adaptation with multiple sources and domain adaptation combining source and target data. This framework is suitable for a variant of learning tasks including classification and regression. Based on the derived bounds, we theoretically analyze the asymptotic convergence and the rate of convergence of the learning process for domain adaptation. There are four important aspects of this framework: the quantity measuring the difference between two domains; the complexity measure of function class, the deviation inequality and the symmetrization inequality for domain adaptation.

\begin{itemize}
  \item We use the integral probability metric $D_{\mathcal{F}}(S,T)$ to measure the difference between two domains $\mathcal{Z}^{(S)}$ and $\mathcal{Z}^{(T)}$. We show that the integral probability metric is well-defined and is a (semi)metric on the space of the probability distributions. It can be bounded by the summation of the {\it discrepancy distance} $\mathrm{disc}_{\ell}(\mathcal{D}^{(S)},\mathcal{D}^{(T)})$ and the quantity $Q^{(T)}_{\mathcal{G}}(g_*^{(S)},g_*^{(T)})$, which measure the difference between the input-space distributions $\mathcal{D}^{(S)}$ and $\mathcal{D}^{(T)}$ and the difference between labeling functions $g_*^{(S)}$ and $g_*^{(T)}$, respectively. Note that there is a special case that is more suitable to the integral probability metric $D_{\mathcal{F}}(S,T)$ than other quantities (see Remark \ref{rem:metric}).
  \item The uniform entropy number and the Rademacher complexity are adopted to achieved the generalization bounds \eqref{eq:crate1.M}; \eqref{eq:crate1.C} and \eqref{eq:RB.Rade}; \eqref{eq:RB.Rade.C}, respectively. It is noteworthy that the generalization bounds \eqref{eq:crate1.M} and \eqref{eq:crate1.C} can lead to the results based on the fat-shattering dimension, respectively \citep[see][Theorem 2.18]{Mendelson03}. According to Theorem 2.6.4 of \citet{Vaart96}, the bounds based on the VC dimension can also be obtained from the results \eqref{eq:crate1.M} and \eqref{eq:crate1.C}, respectively.
  \item Instead of directly applying the classical techniques, we present the specific deviation inequalities for the learning process of domain adaptation. In order to obtain the generalization bounds based on the uniform entropy numbers, we develop the specific Hoeffding-type deviation inequalities for the two types of domain adaptation, respectively (see Appendices \ref{app:Proof1} \& \ref{app:Proof2}). Furthermore, we also generalize the classical McDiarmid's inequality to a more general setting where the independent random variables can take value from different domains (see Appendix \ref{app:Proof3}).
  \item We also develop the related symmetrization inequalities of the learning process for domain adaptation. The derived inequalities incorporate the discrepancy term that is determined by the difference between the source and the target domains and reflects the learning-transfering from the source to the target domains.

\end{itemize}

Based on the derived generalization bounds, we provide a rigorous theoretical analysis of the asymptotic convergence and the rate of convergence of the learning process for either kind of domain adaptation. We also consider the choices of ${\bf w}$ and $\tau$ that affect the rate of convergence of the learning processes for the two types of domain adaptation, respectively. Moreover, we give a comparison
with the previous works \cite{Ben-David10,Crammer06,Crammer08,Mansour08,Mansour09,Blitzer08}
as well as the related results of the learning process under the assumption of same distribution \citep[see][]{Bousquet04,Mendelson03}. The numerical experiments support our theoretical findings as well.

In our future work, we will attempt to find a new distance between distributions to develop the generalization bounds based on other complexity measures, and analyze other theoretical properties of domain adaptation.
%

\newpage

\appendix

\section{Proof of Theorem \ref{thm:crate.M}} \label{app:Proof1}

In this appendix, we provide the proof of Theorem \ref{thm:crate.M}. In order to achieve the proof, we need to develop the specific Hoeffding-type deviation inequality and the symmetrization inequality for domain adaptation with multiple sources.
\subsection{Hoeffding-Type Deviation Inequality for Multiple Sources}

Deviation (or concentration) inequalities play an essential role in obtaining the generalization bounds for a certain learning process. Generally, specific deviation inequalities need to be developed for different learning processes. There are many popular deviation and concentration inequalities, {\it e.g.}, Hoeffding's inequality, McDiarmid's inequality, Bennett's inequality, Bernstein's inequality and Talagrand's inequality. These results are all built under the assumption of same distribution, and thus they are
not applicable (or at least cannot be directly applied) to the setting of multiple sources. Next, based on Hoeffding's inequality \citep{Hoeffding63}, we present a deviation inequality for multiple sources.

\begin{theorem}\label{thm:dineq.M}
Assume that $\mathcal{F}$ is a function class consisting of bounded functions with the range $[a,b]$. Let ${\bf Z}_{1}^{N_k}=\{{\bf z}^{(k)}_n\}_{n=1}^{N_k}$ be the set of i.i.d. samples drawn from the source domain $\mathcal{Z}^{(S_k)}\subset\mathbb{R}^L$ $(1\leq k\leq K)$. Given ${\bf w}\in[0,1]^{K}$ with $\sum_{k=1}^Kw_k=1$ and for any $f\in\mathcal{F}$, we define a function $F_{\bf w}:\mathbb{R}^{L\sum_{k=1}^KN_k}\rightarrow\mathbb{R}$ as
\begin{equation}\label{eq:F.M}
    F_{\bf w}\left(\{{\bf X}_{1}^{N_k}\}_{k=1}^K\right):=\sum_{k=1}^Kw_k
    \Big(\prod_{i\not=k}N_i\Big)\sum_{n=1}^{N_k}f({\bf x}^{(k)}_n),
\end{equation}
where for any $1\leq k\leq K$ and given $N_k\in\mathbb{N}$, the set ${\bf X}_1^{N_k}$ is denoted as $${\bf X}_1^{N_k}:=\{{\bf x}_1^{(k)},{\bf x}_2^{(k)},\cdots,{\bf x}_{N_k}^{(k)}\}\in(\mathbb{R}^{L})^{N_k}.$$
Then, we have for any $\xi>0$,
\begin{align}\label{eq:dineq.M}
&\mathrm{Pr}\left\{\big|\mathrm{E}^{(S)}F_{\bf w}-F_{\bf w}\big(\{{\bf Z}_{1}^{N_k}\}_{k=1}^K\big)\big|>\xi\right\}\nonumber\\
    \leq&2\exp\left\{-\frac{2\xi^2}{(b-a)^2
\big(\prod_{k=1}^KN_k\big)
\Big(\sum_{k=1}^{K}w_k^2\big(\prod_{i\not=k}N_i\big)\Big)}\right\},
\end{align}
where $\mathrm{E}^{(S)}$ stands for the expectation taken on all source domains $\{\mathcal{Z}^{(S_k)}\}_{k=1}^K$.
\end{theorem}

This result is an extension of the classical Hoeffding-type deviation inequality under the assumption of same distribution \citep[see][Theorem 1]{Bousquet04}. Compared to the classical result, the resultant deviation inequality \eqref{eq:dineq.M} is suitable to the setting of multiple sources. These two inequalities coincide when there is only one source, {\it i.e.}, $K=1$

The proof of Theorem \ref{thm:dineq.M} is processed by a martingale method. Before the formal proof, we introduce some essential notations.

Let $\{{\bf Z}_{1}^{N_k}\}_{k=1}^K$ be sample sets drawn from multiple sources $\{\mathcal{Z}^{(S_k)}\}_{k=1}^K$, respectively.
Define a random variable
\begin{equation}\label{eq:Sn1.M}
    S^{(k)}_{n}:=\mathrm{E}^{(S)}\left\{F_{\bf w}(\{{\bf Z}_{1}^{N_k}\}_{k=1}^K)|{\bf Z}_{1}^{N_1},{\bf Z}_{1}^{N_2},\cdots,{\bf Z}_{1}^{N_{k-1}},{\bf Z}_1^{n}\right\},\;\; 1\leq k\leq K,\;0\leq n\leq N_k,
\end{equation}
where
$${\bf Z}_{1}^{n}=\{{\bf z}^{(k)}_1,{\bf z}^{(k)}_2,\cdots,{\bf z}^{(k)}_{n}\}\subseteq{\bf Z}_{1}^{N_k},  \mbox{  and   }   {\bf Z}_{1}^0=\varnothing.$$
It is clear that
$$S_0^{(1)}=\mathrm{E}^{(S)}F_{\bf w} \mbox{  and }  S^{(K)}_{N_K}=F_{\bf w}(\{{\bf Z}_{1}^{N_k}\}_{k=1}^K),$$
where $\mathrm{E}^{(S)}$ stands for the expectation taken on all source domains $\{\mathcal{Z}^{(S_k)}\}_{k=1}^K$.

Then, according to \eqref{eq:F.M} and \eqref{eq:Sn1.M},
we have for any $1\leq k\leq K$ and $1\leq n\leq N_k$,
\begin{align}\label{eq:Sn2.M}
S^{(k)}_n-S^{(k)}_{n-1}
=&\mathrm{E}^{(S)}\left\{F_{\bf w}(\{{\bf Z}_{1}^{N_k}\}_{k=1}^K)\big|{\bf Z}_{1}^{N_1},{\bf Z}_{1}^{N_2},\cdots,{\bf Z}_{1}^{N_{k-1}},{\bf Z}_1^{n}\right\}\nonumber\\
&-\mathrm{E}^{(S)}\left\{F_{\bf w}(\{{\bf Z}_{1}^{N_k}\}_{k=1}^K)\big|{\bf Z}_{1}^{N_1},{\bf Z}_{1}^{N_2},\cdots,{\bf Z}_{1}^{N_{k-1}},{\bf Z}_1^{n-1}\right\}\nonumber\\
    =&\mathrm{E}^{(S)}\left\{\sum_{k=1}^Kw_k
    \Big(\prod_{i\not=k}N_i\Big)\sum_{n=1}^{N_k}f({\bf z}^{(k)}_n)\big|{\bf Z}_{1}^{N_1},{\bf Z}_{1}^{N_2},\cdots,{\bf Z}_{1}^{N_{k-1}},{\bf Z}_1^{n}\right\}\nonumber\\
    & -\mathrm{E}^{(S)}\left\{\sum_{k=1}^Kw_k\Big(\prod_{i\not=k}N_i\Big)
    \sum_{n=1}^{N_k}f({\bf z}^{(k)}_n)\big|{\bf Z}_{1}^{N_1},{\bf Z}_{1}^{N_2},\cdots,{\bf Z}_{1}^{N_{k-1}},{\bf Z}_1^{n-1}\right\}\nonumber\\
     =&\sum_{l=1}^{k-1}w_l\Big(\prod_{i\not=l}N_i\Big)\sum_{j=1}^{N_l}f({\bf z}^{(l)}_j)+w_k\Big(\prod_{i\not=k}N_i\Big)\sum_{j=1}^nf({\bf z}_j^{(k)})\nonumber\\
     &+\mathrm{E}^{(S)}\left\{\sum_{l=k+1}^Kw_l
     \Big(\prod_{i\not=l}N_i\Big)\sum_{j=1}^{N_l}f({\bf z}^{(l)}_j)+w_k\Big(\prod_{i\not=k}N_i\Big)\sum_{j=n+1}^{N_k}f({\bf z}^{(k)}_j)\right\}\nonumber\\
     &-
    \sum_{l=1}^{k-1}w_l\Big(\prod_{i\not=l}N_i\Big)\sum_{j=1}^{N_l}f({\bf z}^{(l)}_j)-w_k\Big(\prod_{i\not=k}N_i\Big)\sum_{j=1}^{n-1}f({\bf z}_j^{(k)})\nonumber\\
     &-\mathrm{E}^{(S)}\left\{\sum_{l=k+1}^Kw_l
     \Big(\prod_{i\not=l}N_i\Big)\sum_{j=1}^{N_l}f({\bf z}^{(l)}_j)+w_k\Big(\prod_{i\not=k}N_i\Big)\sum_{j=n}^{N_k}f({\bf z}^{(k)}_j)\right\}\nonumber\\
    =&w_k\Big(\prod_{i\not=k}N_i\Big)\left(f({\bf z}_n^{(k)})-\mathrm{E}^{(S_k)}f\right).
\end{align}

To prove Theorem \ref{thm:dineq.M}, we need the following inequality resulted from Hoeffding's lemma.

\begin{lemma}\label{lem:Hoeffding}
Let $f$ be a function with the range $[a,b]$. Then, the following holds for any $\alpha>0$:
\begin{align}\label{eq:Hoeffding}
&\mathrm{E}
\left\{\mathrm{e}^{\alpha(f({\bf z}^{(S)})-\mathrm{E}^{(S)}f)}\right\}
\leq \mathrm{e}^{\frac{\alpha^2(b-a)^2}{8}}.
\end{align}
\end{lemma}

{\it Proof.} We consider
$$(f({\bf z}^{(S)})-\mathrm{E}^{(S)}f)$$
as a random variable. Then, it is clear that
$$\mathrm{E}\{f({\bf z}^{(S)})-\mathrm{E}^{(S)}f\}=0.$$
Since the value of $\mathrm{E}^{(S)}f$ is a constant denoted as $e$, we have
$$a-e\leq f({\bf z}^{(S)})-\mathrm{E}^{(S)}f\leq b-e.$$
According to Hoeffding's lemma, we then have
\begin{align}\label{eq:Hoeffding1}
&\mathrm{E}
\left\{\mathrm{e}^{\alpha(f({\bf z}^{(S)})-\mathrm{E}^{(S)}f)}\right\}
\leq \mathrm{e}^{\frac{\alpha^2(b-a)^2}{8}}.
\end{align}
 This completes the proof. \hfill $\blacksquare$

We are now ready to prove Theorem \ref{thm:dineq.M}.

{\it Proof of Theorem \ref{thm:dineq.M}.}
According to \eqref{eq:F.M}, \eqref{eq:Sn2.M}, Lemma \ref{lem:Hoeffding}, Markov's inequality, Jensen's inequality and the law of iterated expectation, we have for any $\alpha>0$,
\begin{align}\label{eq:main1.M}
   &\mathrm{Pr}\left\{F_{\bf w}\big(\{{\bf Z}_{1}^{N_k}\}_{k=1}^K\big)-\mathrm{E}^{(S)}F_{\bf w}>\xi\right\}\nonumber\\
   \leq& \mathrm{e}^{-\alpha\xi}
\mathrm{E}\left\{\mathrm{e}^{\alpha\left(F_{\bf w}\big(\{{\bf Z}_{1}^{N_k}\}_{k=1}^K\big)-\mathrm{E}^{(S)}F_{\bf w}\right)}\right\}\nonumber\\
=&\mathrm{e}^{-\alpha\xi}\mathrm{E}\left\{\mathrm{E}
\left\{\mathrm{e}^{\alpha\sum_{k=1}^K\sum_{n=1}^{N_k}
\left(S^{(k)}_{n}-S^{(k)}_{n-1}\right)}\big|{\bf Z}_{1}^{N_1},\cdots,{\bf Z}_{1}^{N_{K-1}},{\bf Z}_{1}^{N_{K}-1}\right\}\right\}\nonumber\\
=&\mathrm{e}^{-\alpha\xi}\mathrm{E}\left\{\mathrm{e}^
{\alpha\left(\sum_{k=1}^{K}\sum_{n=1}^{N_k}
\left(S^{(k)}_{n}-S^{(k)}_{n-1}\right)-
\left(S^{(K)}_{N_K}-S^{(K)}_{N_K-1}\right)\right)}
\mathrm{E}
\left\{\mathrm{e}^{\alpha\left(S^{(K)}_{N_K}-S^{(K)}_{N_K-1}\right)}
\big|{\bf Z}_{1}^{N_1},\cdots,{\bf Z}_{1}^{N_{K-1}},{\bf Z}_{1}^{N_{K}-1}\right\}\right\}\nonumber\\
=&\mathrm{e}^{-\alpha\xi}\mathrm{E}\left\{\mathrm{e}^
{\alpha\left(\sum_{k=1}^{K}\sum_{n=1}^{N_k}
\left(S^{(k)}_{n}-S^{(k)}_{n-1}\right)-
\left(S^{(K)}_{N_K}-S^{(K)}_{N_K-1}\right)\right)}
\mathrm{E}
\left\{\mathrm{e}^{\alpha w_K(\prod_{i\not=K}N_i)(f({\bf z}_N^{(K)})-\mathrm{E}^{(S_K)}f)}\right\}\right\}\nonumber\\
\leq&\mathrm{e}^{-\alpha\xi}\mathrm{E}\left\{\mathrm{e}^
{\alpha\left(\sum_{k=1}^{K}\sum_{n=1}^{N_k}
\left(S^{(k)}_{n}-S^{(k)}_{n-1}\right)-
\left(S^{(K)}_{N_K}-S^{(K)}_{N_K-1}\right)\right)}\right\}
\mathrm{e}^{\frac{\alpha^2w_K^2(\prod_{i\not=K}N_i)^2(b-a)^2}{8}},
\end{align}
where ${\bf Z}_1^{N_K-1}:=\{{\bf z}_1^{(K)},\cdots,{\bf z}_{N_K-1}^{(K)}\}\subset{\bf Z}_1^{N_K}$.
Therefore, we have
\begin{equation}\label{eq:main2.M}
   \mathrm{Pr}\left\{F_{\bf w}\left(\{{\bf Z}_{1}^{N_k}\}_{k=1}^K\right)-\mathrm{E}^{(S)}F_{\bf w}>\xi\right\}\leq\mathrm{e}^{\Phi(\alpha)-\alpha\xi},
\end{equation}
where
\begin{align}\label{eq:Hn.M}
    \Phi(\alpha)
=&\frac{\alpha^2(b-a)^2\big(\prod_{k=1}^KN_k\big)\Big(\sum_{k=1}^{K}w_k^2
\big(\prod_{i\not=k}N_i\big)\Big)}{8}.
\end{align}
Similarly, we can obtain
\begin{align}\label{eq:deviation3.M}
    \mathrm{Pr}\left\{\mathrm{E}^{(S)}F_{\bf w}-F_{\bf w}\big(\{{\bf Z}_{1}^{N_k}\}_{k=1}^K\big)>\xi\right\} \leq\mathrm{e}^{\Phi(\alpha)-\alpha\xi}.
\end{align}
Note that $\Phi(\alpha)-\alpha\xi$ is a quadratic function with respect to $\alpha>0$ and thus the minimum value ``$\min_{\alpha>0}\left\{\Phi(\alpha)-\alpha\xi\right\}$" is achieved
when
$$\alpha=\frac{4\xi}{(b-a)^2\big(\prod_{k=1}^KN_k\big)
\Big(\sum_{k=1}^{K}w_k^2\big(\prod_{i\not=k}N_i\big)\Big)}.$$
By combining \eqref{eq:main2.M}, \eqref{eq:Hn.M} and \eqref{eq:deviation3.M}, we arrive at
\begin{align*}
    &\mathrm{Pr}\left\{\big|\mathrm{E}^{(S)}F_{\bf w}-F_{\bf w}\big(\{{\bf Z}_{1}^{N_k}\}_{k=1}^K\big)\big|>\xi\right\}\nonumber\\
    \leq&2\exp\left\{-\frac{2\xi^2}{(b-a)^2
\big(\prod_{k=1}^KN_k\big)
\Big(\sum_{k=1}^{K}w_k^2\big(\prod_{i\not=k}N_i\big)\Big)}\right\}.
\end{align*}
This completes the proof.  \hfill $\blacksquare$

In the following subsection, we present a symmetrization inequality for domain adaptation with multiple sources.

\subsection{Symmetrization Inequality}

Symmetrization inequalities are mainly used to replace the expected risk by an empirical risk computed on another sample set that is independent of the given sample set but has the same distribution. In this manner, the generalization bounds can be achieved by applying some kinds of complexity measures, {\it e.g.}, the covering number and the VC dimension. However, the classical symmetrization results are built under the assumption of same distribution \citep[see][]{Bousquet04}. The symmetrization inequality for domain adaptation with multiple sources is presented in the following theorem:

\begin{theorem}\label{thm:sym.M}
Assume that $\mathcal{F}$ is a function class with the range $[a,b]$. Let sample sets
$\{{\bf Z}_1^{N_k}\}_{k=1}^K$ and $\{{\bf Z'}_1^{N_k}\}_{k=1}^K$
be drawn from the source domains $\{\mathcal{Z}^{(S_k)}\}_{k=1}^K$.
Then, given an arbitrary $\xi>D_{\mathcal{F}}^{({\bf w})}(S,T)$ and ${\bf w}=(w_1,\cdots,w_K)\in[0,1]^{K}$ with $\sum_{k=1}^Kw_k=1$, we have for any $\big(\prod_{k=1}^KN_k\big)\geq\frac{8\left(b-a\right)^2}
{(\xi')^2}$,
\begin{align}\label{eq:sym1.M}
    \mathrm{Pr}\left\{\sup_{f\in\mathcal{F}}\big|\mathrm{E}^{(T)}f-\mathrm{E}^{(S)}_{\bf w}f\big|>\xi\right\}
    \leq2\mathrm{Pr}\left\{\sup_{f\in\mathcal{F}}\big|\mathrm{E'}^{(S)}_{\bf w}f-\mathrm{E}^{(S)}_{\bf w}f\big|>\frac{\xi'}{2}\right\},
    \end{align}
    where $\xi'=\xi-D_{\mathcal{F}}^{({\bf w})}(S,T).$
\end{theorem}

This theorem shows that given $\xi>D_{\mathcal{F}}^{({\bf w})}(S,T)$, the probability of the event:
\begin{equation*}
\sup_{f\in\mathcal{F}}\big|\mathrm{E}^{(T)}f-\mathrm{E}^{(S)}_{\bf w}f\big|>\xi
\end{equation*}
can be bounded by using the probability of the event:
\begin{equation}
\sup_{f\in\mathcal{F}}\big|\mathrm{E'}^{(S)}_{\bf w}f-\mathrm{E}^{(S)}_{\bf w}f\big|>\frac{\xi-D_{\mathcal{F}}^{({\bf w})}(S,T)}{2}
\end{equation}
that is only determined by the characteristics of the source domains $\{\mathcal{Z}^{(S_k)}\}_{k=1}^K$ when $\prod_{k=1}^KN_k\geq \frac{8(b-a)^2}{(\xi')^2}$ with $\xi'=\xi-D_{\mathcal{F}}^{({\bf w})}(S,T)$. Compared to the classical symmetrization result under the assumption of same distribution \citep[see][]{Bousquet04}, there is a discrepancy term $D_{\mathcal{F}}^{({\bf w})}(S,T)$ in the derived inequality. Especially, the two results coincide when any source domain and the target domain match, {\it i.e.,} $D_{\mathcal{F}}(S_k,T)=0$ holds for any $1\leq k\leq K$. The following is the proof of Theorem \ref{thm:sym.M}.

{\it Proof of Theorem \ref{thm:sym.M}.}
Let $\widehat{f}$ be the function achieving the supremum: $$\sup_{f\in\mathcal{F}}\big|\mathrm{E}^{(T)}f-\mathrm{E}^{(S)}_{\bf w}f\big|$$ with respect to the sample set $\{{\bf Z}_1^{N_k}\}_{k=1}^K$. According to \eqref{eq:short1.M}, \eqref{eq:short2.M}, \eqref{eq:distance} and \eqref{eq:Dist.M}, we arrive at
\begin{align}\label{eq:sym00.M}
 |\mathrm{E}^{(T)}\widehat{f}-\mathrm{E}^{(S)}_{\bf w}\widehat{f}|=|\mathrm{E}^{(T)}\widehat{f}
 -\overline{\mathrm{E}}^{(S)}\widehat{f}
 +\overline{\mathrm{E}}^{(S)}\widehat{f}-\mathrm{E}^{(S)}_{\bf w}\widehat{f}|
 \leq D_\mathcal{F}^{({\bf w})}(S,T)+\big|\overline{\mathrm{E}}^{(S)}\widehat{f}-\mathrm{E}^{(S)}_{\bf w}\widehat{f}\big|,
\end{align}
and thus,
\begin{align}\label{eq:sym01.M}
 \mathrm{Pr}\left\{|\mathrm{E}^{(T)}\widehat{f}-\mathrm{E}^{(S)}_{\bf w}\widehat{f}|>\xi\right\}
 \leq \mathrm{Pr}\left\{D_\mathcal{F}^{({\bf w})}(S,T)+\big|\overline{\mathrm{E}}^{(S)}\widehat{f}-\mathrm{E}^{(S)}_{\bf w}\widehat{f}\big|>\xi\right\},
\end{align}
where the expectation $\mathrm{E}^{(S)}\widehat{f}$ is defined as
\begin{equation}\label{eq:ESf}
\overline{\mathrm{E}}^{(S)}\widehat{f}:=
\sum_{k=1}^Kw_k\mathrm{E}^{(S_k)}\widehat{f}.
\end{equation}

Let
\begin{align}\label{eq:xi1.M}
 \xi':=\xi-D_\mathcal{F}^{({\bf w})}(S,T),
\end{align}
and denote $\wedge$ as the conjunction of two events. According to the triangle inequality, we have
\begin{equation*}
\Big(|\overline{\mathrm{E}}^{(S)}\widehat{f}-\mathrm{E}^{(S)}_{\bf w}\widehat{f}|-|\mathrm{E'}^{(S)}_{\bf w}\widehat{f}-\overline{\mathrm{E}}^{(S)}\widehat{f}| \Big)\leq
|\mathrm{E'}^{(S)}_{\bf w}\widehat{f}-\mathrm{E}^{(S)}_{\bf w}\widehat{f}|,
\end{equation*}
and thus for any $\xi'>0$,
\begin{align*}
 &\left(\mathbf{1}_{|\overline{\mathrm{E}}^{(S)}\widehat{f}-\mathrm{E}^{(S)}_{\bf w}\widehat{f}|>\xi'}\right)
\left(\mathbf{1}_{|\overline{\mathrm{E}}^{(S)}\widehat{f}-\mathrm{E'}^{(S)}_{\bf w}\widehat{f}|<\frac{\xi'}{2}}\right)\nonumber\\
=&\mathbf{1}_{\left\{|\overline{\mathrm{E}}^{(S)}\widehat{f}-\mathrm{E}^{(S)}_{\bf w}\widehat{f}|>\xi'\right\}
\,\wedge\,\left\{|\overline{\mathrm{E}}^{(S)}\widehat{f}-\mathrm{E'}^{(S)}_{\bf w}\widehat{f}|<\frac{\xi'}{2}\right\}}\nonumber\\
\leq&\mathbf{1}_{|\mathrm{E'}^{(S)}_{\bf w}\widehat{f}-\mathrm{E}^{(S)}_{\bf w}\widehat{f}|>\frac{\xi'}{2}}\;.
\end{align*}
Then, taking the expectation with respect to $\{{\bf Z'}_1^{N_k}\}_{k=1}^K$ gives
\begin{align}\label{eq:the:sym.pr2.M}
   &\left(\mathbf{1}_{|\overline{\mathrm{E}}^{(S)}\widehat{f}-\mathrm{E}^{(S)}_{\bf w}\widehat{f}|>\xi'}\right)
\mathrm{Pr'}\left\{|\overline{\mathrm{E}}^{(S)}\widehat{f}-\mathrm{E'}^{(S)}_{\bf w}\widehat{f}|<\frac{\xi'}{2}\right\}\nonumber\\
\leq& \mathrm{Pr'}\left\{|\mathrm{E'}^{(S)}_{\bf w}\widehat{f}-\mathrm{E}^{(S)}_{\bf w}\widehat{f}|>\frac{\xi'}{2}\right\}.
\end{align}

By Chebyshev's inequality, since $\{{\bf Z'}_1^{N_k}\}_{k=1}^K$ are the sets of i.i.d. samples drawn from the multiple sources $\{\mathcal{Z}^{(S_k)}\}_{k=1}^K$ respectively, we have for any $\xi'>0$,
\begin{align}\label{eq:the:sym.pr3.M}
   \mathrm{Pr'}\left\{\big|\overline{\mathrm{E}}^{(S)}\widehat{f}-\mathrm{E'}^{(S)}_{\bf w}\widehat{f}\,\big|\geq\frac{\xi'}{2}\right\}
   \leq&\mathrm{Pr'}\left\{\sum_{k=1}^K\frac{w_k}{N_k}
   \sum_{n=1}^{N_k}|\mathrm{E}^{(S_k)}\widehat{f}-\widehat{f}({\bf z'}_{n}^{(k)})|\geq\frac{\xi'}{2}\right\}\nonumber\\
   =&\mathrm{Pr'}\left\{\sum_{k=1}^Kw_k
   \Big(\prod_{i\not=k}N_i\Big)\sum_{n=1}^{N_k}|\mathrm{E}^{(S_k)}\widehat{f}-\widehat{f}({\bf z'}_{n}^{(k)})|\geq\frac{\xi'\prod_{k=1}^KN_k}{2}\right\}\nonumber\\
\leq& \frac{4\mathrm{E}\left\{\sum_{k=1}^Kw_k\big(\prod_{i\not=k}N_i\big)
\sum_{n=1}^{N_k}\big|\mathrm{E}^{(S_k)}\widehat{f}-\widehat{f}({\bf z'}_n^{(k)})\big|^2\right\}}{\big(\prod_{k=1}^KN_k\big)^2(\xi')^2}\nonumber\\
=& \frac{4\mathrm{E}\left\{\sum_{k=1}^Kw_k\big(\prod_{i\not=k}N_i\big)
N_k\left(b-a\right)^2\right\}}{\big(\prod_{k=1}^KN_k\big)^2(\xi')^2}\nonumber\\
=& \frac{4\big(\prod_{k=1}^KN_k\big)\left(b-a\right)^2}
{\big(\prod_{k=1}^KN_k\big)^2(\xi')^2}
= \frac{4\left(b-a\right)^2}{(\xi')^2\big(\prod_{k=1}^KN_k\big)}.
\end{align}

Subsequently, according to \eqref{eq:the:sym.pr2.M} and \eqref{eq:the:sym.pr3.M}, we have for any $\xi'>0$,
\begin{align}\label{eq:the:sym.pr4.M}
   \mathrm{Pr'}\left\{|\mathrm{E'}^{(S)}_{\bf w}\widehat{f}-\mathrm{E}^{(S)}_{\bf w}\widehat{f}|>\frac{\xi'}{2}\right\}
\geq
   \left(\mathbf{1}_{|\overline{\mathrm{E}}^{(S)}\widehat{f}-\mathrm{E}^{(S)}_{\bf w}\widehat{f}|>\xi'}\right)\left(1-\frac{4\left(b-a\right)^2}
   {(\xi')^2\big(\prod_{k=1}^KN_k\big)}\right).
\end{align}
By combining \eqref{eq:sym01.M}, \eqref{eq:xi1.M} and \eqref{eq:the:sym.pr4.M}, taking the expectation with respect to $\{{\bf Z}_1^{N_k}\}_{k=1}^K$ and letting
$$\frac{4\left(b-a\right)^2}{(\xi')^2\big(\prod_{k=1}^KN_k\big)}\leq \frac{1}{2}$$
can lead to: for any $\xi>D_\mathcal{F}^{({\bf w})}(S,T)$,
\begin{align}\label{eq:sym.result1.M}
  \mathrm{Pr}\left\{|\mathrm{E}^{(T)}\widehat{f}-\mathrm{E}^{(S)}_{\bf w}\widehat{f}|>\xi\right\}
 \leq& \mathrm{Pr}\left\{|\overline{\mathrm{E}}^{(S)}\widehat{f}-\mathrm{E}^{(S)}_{\bf w}\widehat{f}|>\xi'\right\}\nonumber\\
 \leq& 2\mathrm{Pr}\left\{|\mathrm{E'}^{(S)}_{\bf w}\widehat{f}-\mathrm{E}^{(S)}_{\bf w}\widehat{f}|>\frac{\xi'}{2}\right\}
\end{align}
with $\xi'=\xi-D_\mathcal{F}^{({\bf w})}(S,T)$.
This completes the proof.
\hfill$\blacksquare$

By using the resultant deviation inequality and the symmetrization inequality, we can achieve the proof of Theorem \ref{thm:crate.M}.

%
\subsection{Proof of Theorem \ref{thm:crate.M}}

{\it Proof of Theorem \ref{thm:crate.M}.} Consider $\epsilon$ as an independent Rademacher random variables, {\it i.e.}, an independent $\{-1,1\}$-valued random variable with equal probability of taking either value. Given sample sets $\{{\bf Z}_{1}^{2N_k}\}_{k=1}^K$, denote for any $f\in\mathcal{F}$ and $1\leq k\leq K$,
\begin{equation}\label{eq:epsilon.M}
    \overrightarrow{\epsilon}^{(k)}:=(\epsilon^{(k)}_1,\cdots,
    \epsilon^{(k)}_{N_k},
    -\epsilon^{(k)}_1,\cdots,-\epsilon^{(k)}_{N_k})
    \in\{-1,1\}^{2N_k},
\end{equation}
and for any $f\in\mathcal{F}$,
\begin{equation}\label{eq:vecf.M}
  \overrightarrow{f}({\bf Z}_{1}^{2N_k}):=\big(f({\bf z'}^{(k)}_1),\cdots,f({\bf z'}^{(k)}_{N_k}),f({\bf z}^{(k)}_1),\cdots,f({\bf z}^{(k)}_{N_k})\big).
\end{equation}

According to \eqref{eq:short1.M}, \eqref{eq:short2.M} and Theorem \ref{thm:sym.M}, given an arbitrary $\xi>D_\mathcal{F}^{({\bf w})}(S,T)$, we have for any $\{N_k\}_{k=1}^K\in\mathbb{N}^K$ such that $\prod_{k=1}^KN_k\geq\frac{8(b-a)^2}{(\xi')^2}$ with $\xi'=\xi-D_\mathcal{F}^{({\bf w})}(S,T)$,
\begin{align}\label{eq:bas1.M}
    &\mathrm{Pr}\left\{\sup_{f\in\mathcal{F}}
    \big|\mathrm{E}^{(T)}f-\mathrm{E}^{(S)}_{\bf w}f\big|>\xi\right\}\nonumber\\
\leq& 2\mathrm{Pr}\left\{\sup_{f\in\mathcal{F}}
\big|\mathrm{E'}^{(S)}_{\bf w}f-\mathrm{E}_{\bf w}^{(S)}f\big|>\frac{\xi'}{2}\right\}\qquad\mbox{(by Theorem \ref{thm:sym.M})}\nonumber\\
=&2\mathrm{Pr}\left\{\sup_{f\in\mathcal{F}}
\Big|\sum_{k=1}^K\frac{w_k}{N_k}\sum_{n=1}^{N_k}\big(f({\bf z'}^{(k)}_n)-f({\bf z}^{(k)}_n)\big)\Big|>\frac{\xi'}{2}\right\}\nonumber\\
=&2\mathrm{Pr}\left\{\sup_{f\in\mathcal{F}}
\Big|\sum_{k=1}^K\frac{w_k}{N_k}\sum_{n=1}^{N_k}
\epsilon^{(k)}_n\big(f({\bf z'}^{(k)}_n)-f({\bf z}^{(k)}_n)\big)\Big|>\frac{\xi'}{2}\right\}\nonumber\\
=&2\mathrm{Pr}\left\{\sup_{f\in\mathcal{F}}
\Big|\sum_{k=1}^K\frac{w_k}{2N_k}
\big\langle\overrightarrow{\epsilon}^{(k)},
\overrightarrow{f}({\bf Z}_{1}^{2N_k})\big\rangle\Big|>\frac{\xi'}{4}\right\}.
\qquad\mbox{(by \eqref{eq:epsilon.M} and \eqref{eq:vecf.M})}
\end{align}

Fix a realization of $\{{\bf Z}_{1}^{2N_k}\}_{k=1}^K$ and let $\Lambda$ be a $\xi'/8$-radius cover of $\mathcal{F}$ with respect to the $\ell_1^{\bf w}(\{{\bf Z}_{1}^{2N_k}\}_{k=1}^K)$ norm.
Since $\mathcal{F}$ is composed of the bounded functions with the range $[a,b]$, we assume that the same holds for any $h\in\Lambda$. If $f_0$ is the function that achieves the following supremum
$$\sup_{f\in\mathcal{F}}\Big|\sum_{k=1}^K\frac{w_k}{2N_k}
\big\langle\overrightarrow{\epsilon}^{(k)},\overrightarrow{f}({\bf Z}_{1}^{2N_k})\big\rangle\Big|>\frac{\xi'}{4},$$
there must be an $h_0\in\Lambda$ that satisfies
\begin{equation*}
    \sum_{k=1}^K\frac{w_k}{2N_k}\left(|f_0({\bf z'}^{(k)}_n)-h_0({\bf z'}^{(k)}_n)|+|f_0({\bf z}^{(k)}_n)-h_0({\bf z}^{(k)}_n)|\right)< \frac{\xi'}{8},
\end{equation*}
and meanwhile,
\begin{equation*}
    \Big|\sum_{k=1}^K\frac{w_k}{2N_k}
    \big\langle\overrightarrow{\epsilon}^{(k)},
    \overrightarrow{h_0}({\bf Z}_{1}^{2N_k})\big\rangle\Big|>\frac{\xi'}{8}.
\end{equation*}
Therefore, for the realization of $\{{\bf Z}_{1}^{2N_k}\}_{k=1}^K$, we arrive at
\begin{align}\label{eq:bas2.M}
&\mathrm{Pr}\left\{\sup_{f\in\mathcal{F}}
\Big|\sum_{k=1}^K\frac{w_k}{2N_k}
\big\langle\overrightarrow{\epsilon}^{(k)},
\overrightarrow{f}({\bf Z}_{1}^{2N_k})\big\rangle\Big|>\frac{\xi'}{4}\right\}\nonumber\\
\leq&\mathrm{Pr}\left\{\sup_{h\in\Lambda}
\Big|\sum_{k=1}^K\frac{w_k}{2N_k}
\big\langle\overrightarrow{\epsilon}^{(k)},
\overrightarrow{h}({\bf Z}_{1}^{2N_k})\big\rangle\Big|>\frac{\xi'}{8}\right\}.
\end{align}

Moreover, we denote the event
\begin{equation*}
A:=\left\{\mathrm{Pr}\left\{\sup_{h\in\Lambda}
\Big|\sum_{k=1}^K\frac{w_k}{2N_k}
\big\langle\overrightarrow{\epsilon}^{(k)},\overrightarrow{h}({\bf Z}_{1}^{2N_k})\big\rangle\Big|>\frac{\xi'}{8}\right\}\right\},
\end{equation*}
and let ${\bf 1}_A$ be the characteristic function of the event $A$. By Fubini's Theorem, we have
\begin{align}\label{eq:Fubini.M}
&\mathrm{Pr}\{A\}=\mathrm{E}
\Big\{\mathrm{E}_{\overrightarrow{\epsilon}}\big\{{\bf 1}_A\big\}\big|\;\{{\bf Z}_{1}^{2N_k}\}_{k=1}^K\Big\}\nonumber\\
&=\mathrm{E}\left\{\mathrm{Pr}\left\{\sup_{h\in\Lambda}
\Big|\sum_{k=1}^K\frac{w_k}{2N_k}
\big\langle\overrightarrow{\epsilon}^{(k)},\overrightarrow{h}({\bf Z}_{1}^{2N_k})\big\rangle\Big|>\frac{\xi'}{8}\right\}\Big|\;\{{\bf Z}_{1}^{2N_k}\}_{k=1}^K\right\}.
\end{align}

Fix a realization of $\{{\bf Z}_{1}^{2N_k}\}_{k=1}^K$ again. According to \eqref{eq:epsilon.M}, \eqref{eq:vecf.M} and Theorem \ref{thm:dineq.M}, we have
\begin{align}\label{eq:bas3.M}
&\mathrm{Pr}\left\{\sup_{h\in\Lambda}
\Big|\sum_{k=1}^K\frac{w_k}{2N_k}\big\langle
\overrightarrow{\epsilon}^{(k)},\overrightarrow{h}({\bf Z}_{1}^{2N_k})\big\rangle\Big|>\frac{\xi'}{8}\right\}\nonumber\\
\leq&|\Lambda|\max_{h\in\Lambda}\mathrm{Pr}\left\{
\Big|\sum_{k=1}^K\frac{w_k}{2N_k}
\big\langle\overrightarrow{\epsilon}^{(k)},\overrightarrow{h}({\bf Z}_{1}^{2N_k})\big\rangle\Big|>\frac{\xi'}{8}\right\}\nonumber\\
=&\mathcal{N}\left(\mathcal{F},\xi'/8,\ell^{\bf w}_1(\{{\bf Z}_{1}^{2N_k}\}_{k=1}^K)\right)\max_{h\in\Lambda}
\mathrm{Pr}\left\{\big|\mathrm{E'}^{(S)}_{\bf w}h-\mathrm{E}^{(S)}_{\bf w}h\big|>
\frac{\xi'}{4}\right\}\nonumber\\
\leq& \mathcal{N}\left(\mathcal{F},\xi'/8,\ell^{\bf w}_1(\{{\bf Z}_{1}^{2N_k}\}_{k=1}^K)\right)\max_{h\in\Lambda}\mathrm{Pr}
\left\{|\overline{\mathrm{E}}^{(S)}h-\mathrm{E'}^{(S)}_{\bf w}h|
+|\overline{\mathrm{E}}^{(S)}h-\mathrm{E}^{(S)}_{\bf w}h|
>\frac{\xi'}{4}\right\}\nonumber\\
\leq&
2\mathcal{N}\left(\mathcal{F},\xi'/8,\ell^{\bf w}_1(\{{\bf Z}_{1}^{2N_k}\}_{k=1}^K)\right)\max_{h\in\Lambda}
\mathrm{Pr}\left\{\big|\overline{\mathrm{E}}^{(S)}h-\mathrm{E}^{(S)}_{\bf w}h\big|>\frac{\xi'}{8}\right\}\nonumber\\
\leq&4\mathcal{N}\left(\mathcal{F},\xi'/8,\ell^{\bf w}_1(\{{\bf Z}_{1}^{2N_k}\}_{k=1}^K)\right)
\exp\left\{-\frac{\big(\prod_{k=1}^KN_k\big)
\left(\xi-D^{({\bf w})}_{\mathcal{F}}(S,T)\right)^2}
{32(b-a)^2\big(\sum_{k=1}^{K}w_k^2(\prod_{i\not=k}N_i)\big)}\right\},
\end{align}
where the expectation $\overline{\mathrm{E}}^{(S)}$ is defined in \eqref{eq:ESf}.

The combination of \eqref{eq:bas1.M}, \eqref{eq:bas2.M} and \eqref{eq:bas3.M} leads to the result:  given an arbitrary $\xi>D^{({\bf w})}_{\mathcal{F}}(S,T)$ and for any $\prod_{k=1}^KN_k\geq \frac{8\left(b-a\right)^2}{(\xi')^2}$ with $\xi'=\xi-D^{({\bf w})}_{\mathcal{F}}(S,T)$,
\begin{align}\label{eq:RB1_1.M}
    &\mathrm{Pr}\left\{\sup_{f\in\mathcal{F}}
    \big|\mathrm{E}^{(T)}f-\mathrm{E}^{(S)}_{\bf w}f\big|>\xi\right\}\nonumber\\
        \leq&8\mathrm{E}\mathcal{N}\left(\mathcal{F},\xi'/8,\ell^{\bf w}_1(\{{\bf Z}_{1}^{2N_k}\}_{k=1}^K)\right)
        \exp\left\{-\frac{\big(\prod_{k=1}^KN_k\big)
        \left(\xi-D^{({\bf w})}_{\mathcal{F}}(S,T)\right)^2}{32(b-a)^2
        \big(\sum_{k=1}^{K}w_k^2(\prod_{i\not=k}N_i)\big)}\right\}\nonumber\\
        \leq&8\mathcal{N}^{\bf w}_1\left(\mathcal{F},\xi'/8,2\sum_{k=1}^KN_k\right)
        \exp\left\{-\frac{\big(\prod_{k=1}^KN_k\big)
        \left(\xi-D^{({\bf w})}_{\mathcal{F}}(S,T)\right)^2}{32(b-a)^2
        \big(\sum_{k=1}^{K}w_k^2(\prod_{i\not=k}N_i)\big)}\right\}.
        \end{align}
According to \eqref{eq:RB1_1.M}, letting
\begin{align*}
    \epsilon:=8\mathcal{N}^{\bf w}_1\left(\mathcal{F},\xi'/8,2\sum_{k=1}^KN_k\right)    \exp\left\{-\frac{\big(\prod_{k=1}^KN_k\big)
        \left(\xi-D^{({\bf w})}_{\mathcal{F}}(S,T)\right)^2}{32(b-a)^2
        \big(\sum_{k=1}^{K}w_k^2(\prod_{i\not=k}N_i)\big)}\right\},
\end{align*}
we then arrive at with probability at least $1-\epsilon$,
\begin{align*}
    \sup_{f\in\mathcal{F}}
\big|\mathrm{E}^{(S)}_{\bf w}f-\mathrm{E}^{(T)}f\big|
\leq D^{({\bf w})}_{\mathcal{F}}(S,T)+ \left(\frac{\ln\mathcal{N}^{\bf w}_1\big(\mathcal{F},\xi'/8,2\sum_{k=1}^KN_k\big)-\ln(\epsilon/8)}
{\frac{\big(\prod_{k=1}^KN_k\big)}{32(b-a)^2\big(\sum_{k=1}^{K}w_k^2
(\prod_{i\not=k}N_i)\big)}}\right)^{\frac{1}{2}},
\end{align*}
where $\xi'=\xi-D^{({\bf w})}_{\mathcal{F}}(S,T)$.
This completes the proof. \hfill$\blacksquare$


\section{Proof of Theorem \ref{thm:crate.C}} \label{app:Proof2}

Here, we provide the proof of Theorem \ref{thm:crate.C}. Similar to the situation of domain adaptation with multiple sources, we need to develop the related Hoeffding-type deviation inequality and the symmetrization inequality for domain adaptation combining source and target data.

\subsection{Hoeffding-Type Deviation Inequality}

Based on Hoeffding's inequality \citep{Hoeffding63}, we derive a deviation inequality for the combination of the source and the target domains.

\begin{theorem}\label{thm:dineq.C}
Assume that $\mathcal{F}$ is a function class consisting of bounded functions with the range $[a,b]$. Let ${\bf Z}_{1}^{N_{S}}:=\{{\bf z}_n^{(S)}\}_{n=1}^{N_S}$ and ${\bf \overline{Z}}_{1}^{N_{T}}:=\{{\bf z}_n^{(T)}\}_{n=1}^{N_T}$ be sets of i.i.d. samples drawn from the source domain $\mathcal{Z}^{(S)}\subset\mathbb{R}^L$ and the target domain $\mathcal{Z}^{(T)}\subset\mathbb{R}^L$, respectively. For any $\tau\in[0,1)$,  define a function $F_\tau:\mathbb{R}^{L(N_S+N_T)}\rightarrow\mathbb{R}$ as
\begin{equation}\label{eq:F.C}
    F_\tau\left({\bf X}_{1}^{N_T},{\bf Y}_{1}^{N_S}\right):=\tau N_S\sum_{n=1}^{N_T}f({\bf x}_n)+(1-\tau)N_T\sum_{n=1}^{N_S}f({\bf y}_n),
 \end{equation}
 where
\begin{align*}
   {\bf X}_{1}^{N_T}:=\{{\bf x}_1,\cdots,{\bf x}_{N_T}\}\in(\mathbb{R}^L)^{N_T};\;\;
   {\bf Y}_{1}^{N_S}:=\{{\bf y}_1,\cdots,{\bf y}_{N_S}\}\in(\mathbb{R}^L)^{N_S}.
\end{align*}
Then, we have for any $\tau\in[0,1)$ and any $\xi>0$,
\begin{align}\label{eq:dineq.C}
&\mathrm{Pr}\left\{\big|F_\tau\left({\bf Z}_{1}^{N_S},{\bf\overline{Z}}_{1}^{N_T}\right)
-\mathrm{E}^{(*)}F_\tau\big|>\xi\right\}\nonumber\\
\leq&2\exp\left\{-\frac{2\xi^2}{(b-a)^2N_SN_T
\left((1-\tau)^2N_T+\tau^2N_S\right)}\right\},
\end{align}
where the expectation $\mathrm{E}^{(*)}$ is taken on both of the source domain $\mathcal{Z}^{(S)}$ and the target domain $\mathcal{Z}^{(T)}$.
\end{theorem}

In this theorem, we present a deviation inequality for the combination of source and target domains, which is an extension of the classical Hoeffding-type deviation inequality under the assumption of same distribution \citep[see][Theorem 1]{Bousquet04}. Compare to the classical result, the resultant deviation inequality \eqref{eq:dineq.C} allows the random variables to take values from different domains. The two inequalities coincide when the source domain $\mathcal{Z}^{(S)}$ and the
target domain $\mathcal{Z}^{(T)}$ match, {\it i.e.,} $D_{\mathcal{F}}(S,T)=0$.

The proof of Theorem \ref{thm:dineq.C} is also processed by a martingale method. Before the formal proof, we introduce some essential notations.

For any $\tau\in[0,1)$, we denote
\begin{align}\label{eq:F1.C}
    F_S({\bf Z}_{1}^{N_S}):=(1-\tau)N_T\sum_{n=1}^{N_S}f({\bf z}^{(S)}_n);\;\;
    F_T({\bf \overline{Z}}_{1}^{N_T}):=\tau N_S\sum_{n=1}^{N_T}f({\bf z}^{(T)}_n).
\end{align}
Recalling \eqref{eq:F.C}, it is evident that $F_\tau\big({\bf Z}_{1}^{N_S},{\bf\overline{Z}}_{1}^{N_T}\big)=F_S({\bf Z}_{1}^{N_S})+F_T({\bf \overline{Z}}_{1}^{N_T})$.
We then define two random variables:
\begin{align}\label{eq:Sn1.C}
    S_n:=&\mathrm{E}^{(S)}\left\{F_S({\bf Z}_{1}^{N_S})|{\bf Z}_1^n\right\},\; \mbox{$0\leq n\leq N_S$;}\nonumber \\
    T_n:=&\mathrm{E}^{(T)}\left\{F_T({\bf \overline{Z}}_{1}^{N_T})|{\bf \overline{Z}}_1^n\right\},\; \mbox{$0\leq n\leq N_T$,}
\end{align}
where
\begin{align*}
{\bf Z}_{1}^n&=\{{\bf z}^{(S)}_1,\cdots,{\bf z}^{(S)}_n\}\subseteq{\bf Z}_{1}^{N_{S}} \; \mbox{  with   } \;  {\bf Z}_{1}^0:=\varnothing;\\
{\bf\overline{Z}}_{1}^n&=\{{\bf z}^{(T)}_1,\cdots,{\bf z}^{(T)}_n\}\subseteq{\bf \overline{Z}}_{1}^{N_{T}} \; \mbox{  with   } \;  {\bf \overline{Z}}_{1}^0:=\varnothing.
\end{align*}
It is clear that $S_0=\mathrm{E}^{(S)}F_S$; $S_{N_S}=F_S({\bf Z}_{1}^{N_S})$ and $T_0=\mathrm{E}^{(T)}F_T$;   $T_{N_T}=F_T({\bf\overline{Z}}_{1}^{N_T})$.

According to \eqref{eq:F.C} and \eqref{eq:Sn1.C},
we have for any $1\leq n\leq N_S$ and any $\tau\in[0,1)$,
\begin{align}\label{eq:Sn2.C}
&\,S_n-S_{n-1}\nonumber\\
=&\,\mathrm{E}^{(S)}\left\{F_S({\bf Z}_{1}^{N_S})|{\bf Z}_{1}^n\right\}-\mathrm{E}^{(S)}\left\{F_S({\bf Z}_{1}^{N_S})|{\bf Z}_{1}^{n-1}\right\}\nonumber\\
=&\,\mathrm{E}^{(S)}\left\{(1-\tau)N_T\sum_{n=1}^{N_S}f({\bf z}^{(S)}_n)\Big|{\bf Z}_{1}^n\right\} -\mathrm{E}^{(S)}\left\{(1-\tau)N_T\sum_{n=1}^{N_S}f({\bf z}^{(S)}_n)\Big|{\bf Z}_{1}^{n-1}\right\}\nonumber\\
=&\,(1-\tau)N_T\sum_{m=1}^nf({\bf z}^{(S)}_m)+\mathrm{E}^{(S)}\left\{(1-\tau)N_T\sum_{m=n+1}^{N_S}
f({\bf z}^{(S)}_m)\right\}\nonumber\\
&\,-\left((1-\tau)N_T\sum_{m=1}^{n-1}f({\bf z}^{(S)}_m)+\mathrm{E}^{(S)}\left\{(1-\tau)N_T
\sum_{m=n}^{N_S}f({\bf z}^{(S)}_m)\right\}\right)\nonumber\\
=&\,(1-\tau)N_T\left( f({\bf z}^{(S)}_n)-\mathrm{E}^{(S)}f\right).
\end{align}

Similarly, we also have for any $1\leq n\leq N_T$,
\begin{align}\label{eq:Sn3.C}
T_n-T_{n-1}
=\tau N_S\left(f({\bf z}_n^{(T)})-\mathrm{E}^{(T)}f\right).
\end{align}


We are now ready to prove Theorem \ref{thm:dineq.C}.

{\it Proof of Theorem \ref{thm:dineq.C}.} According to \eqref{eq:F.C} and \eqref{eq:F1.C}, we have
\begin{align}\label{eq:main0.C}
    F_\tau({\bf Z}_1^N)-\mathrm{E}^{(*)}F_\tau
    =&F_S({\bf Z}_{1}^{N_S})+F_T({\bf \overline{Z}}_{1}^{N_T})-
    \mathrm{E}^{(*)}\{F_S+F_T\}\nonumber\\
    =& F_S({\bf Z}_{1}^{N_S})-\mathrm{E}^{(S)}F_S+F_T({\bf \overline{Z}}_{1}^{N_T})-\mathrm{E}^{(T)}F_T.
\end{align}

According to Lemma \ref{lem:Hoeffding}, \eqref{eq:Sn2.C}, \eqref{eq:Sn3.C}, \eqref{eq:main0.C}, Markov's inequality, Jensen's inequality and the law of iterated expectation, we have for any $\alpha>0$ and any $\tau \in[0,1)$,
\begin{align}\label{eq:main1.C}
   &\mathrm{Pr}\left\{F_\tau({\bf Z}_1^N)-\mathrm{E}^{(*)}F_\tau>\xi\right\}\nonumber\\
 =  &\mathrm{Pr}\left\{F_S({\bf Z}_{1}^{N_S})-\mathrm{E}^{(S)}F_S+F_T({\bf \overline{Z}}_{1}^{N_T})-\mathrm{E}^{(T)}F_T>\xi\right\}\nonumber\\
   \leq& \mathrm{e}^{-\alpha\xi}
\mathrm{E}\left\{\mathrm{e}^{\alpha\left(F_S({\bf Z}_{1}^{N_S})-\mathrm{E}^{(S)}F_S+F_T({\bf \overline{Z}}_{1}^{N_T})-\mathrm{E}^{(T)}F_T\right)}\right\}\nonumber\\
=&\mathrm{e}^{-\alpha\xi}
\mathrm{E}\left\{\mathrm{e}^{\alpha\big(\sum_{n=1}^{N_S}(S_n-S_{n-1})
+\sum_{n=1}^{N_T}(T_n-T_{n-1})\big)}\right\}\nonumber\\
=&\mathrm{e}^{-\alpha\xi}\mathrm{E}\left\{\mathrm{E}
\left\{\mathrm{e}^{\alpha\big(\sum_{n=1}^{N_S}(S_n-S_{n-1})
+\sum_{n=1}^{N_T}(T_n-T_{n-1})\big)}\Big|{\bf Z}_1^{N_S-1}\right\}\right\}\nonumber\\
=&\mathrm{e}^{-\alpha\xi}\mathrm{E}
\left\{\mathrm{e}^{\alpha\big(\sum_{n=1}^{N_S-1}(S_n-S_{n-1})
+\sum_{n=1}^{N_T}(T_n-T_{n-1})\big)}
\mathrm{E}
\left\{\mathrm{e}^{\alpha\left(S_{N_S}-S_{N_S-1}\right)}\Big|{\bf Z}_1^{N_S-1}\right\}\right\}\nonumber\\
\leq&\mathrm{e}^{-\alpha\xi}\mathrm{E}
\left\{\mathrm{e}^{\alpha\big(\sum_{n=1}^{N_S-1}(S_n-S_{n-1})
+\sum_{n=1}^{N_T}(T_n-T_{n-1})\big)}\right\}
\mathrm{e}^{\frac{(1-\tau)^2N_T^2\alpha^2(b-a)^2}{8}}\nonumber\\
=&\mathrm{e}^{-\alpha\xi}\mathrm{E}
\left\{\mathrm{e}^{\alpha\big(\sum_{n=1}^{N_S-1}(S_n-S_{n-1})
+\sum_{n=1}^{N_T-1}(T_n-T_{n-1})\big)}
\mathrm{E}
\left\{\mathrm{e}^{\alpha\left(T_{N_T}-T_{N_T-1}\right)}\Big|{\bf \overline{Z}}_1^{N_T-1}\right\}\right\}\nonumber\\
&\times\mathrm{e}^{\frac{(1-\tau)^2N_T^2\alpha^2(b-a)^2}{8}}
\nonumber\\
\leq&\mathrm{e}^{-\alpha\xi}\mathrm{E}
\left\{\mathrm{e}^{\alpha\big(\sum_{n=1}^{N_S-1}(S_n-S_{n-1})
+\sum_{n=1}^{N_T-1}(T_n-T_{n-1})\big)}\right\}
\mathrm{e}^{\frac{\tau^2N_S^2\alpha^2(b-a)^2}{8}}\mathrm{e}^{\frac{(1-\tau)^2N_T^2\alpha^2(b-a)^2}{8}}.
\end{align}
Then, we have
\begin{equation}\label{eq:main2.C}
   \mathrm{Pr}\left\{F_\tau\left({\bf Z}_1^{N_S},{\bf \overline{Z}}_1^{N_T}\right)-\mathrm{E}^{(*)}F_\tau>\xi\right\}
   \leq\mathrm{e}^{\Phi(\alpha)-\alpha\xi},
\end{equation}
where
\begin{align}\label{eq:Hn.C}
    \Phi(\alpha)
=&\frac{\alpha^2(1-\tau)^2(b-a)^2N_SN_T^2}{8}
+\frac{\alpha^2\tau^2(b-a)^2N_S^2N_T}{8}.
\end{align}
Similarly, we can arrive at
\begin{align}\label{eq:deviation3.C}
    \mathrm{Pr}\left\{\mathrm{E}^{(*)}F_\tau-F_\tau\left({\bf Z}_{1}^{N_S},{\bf \overline{Z}}_{1}^{N_T}\right)>\xi\right\} \leq\mathrm{e}^{\Phi(\alpha)-\alpha\xi}.
\end{align}
Note that $\Phi(\alpha)-\alpha\xi$ is a quadratic function with respect to $\alpha>0$ and thus the minimum value
$$\min_{\alpha>0}\left\{\Phi(\alpha)-\alpha\xi\right\}$$
 is achieved
when
$$\alpha=\frac{4\xi}
{(b-a)^2N_SN_T\left((1-\tau)^2N_T+\tau^2N_S\right)}.$$
By combining \eqref{eq:main2.C}, \eqref{eq:Hn.C} and \eqref{eq:deviation3.C}, we arrive at
\begin{align*}
    \mathrm{Pr}\left\{|F_\tau\big({\bf Z}_{1}^{N_S},{\bf \overline{Z}}_{1}^{N_T}\big)-\mathrm{E}^{(*)}F_\tau|>\xi\right\}
    \leq2\exp\left\{-\frac{2\xi^2}
    {(b-a)^2N_SN_T\left((1-\tau)^2N_T+\tau^2N_S\right)}\right\}.
\end{align*}
This completes the proof.  \hfill $\blacksquare$

\subsection{Symmetrization Inequality}

In the following theorem, we present the symmetrization inequality for domain adaptation combining source and target data.

\begin{theorem}\label{thm:sym.C}
Assume that $\mathcal{F}$ is a function class with the range $[a,b]$. Let
${\bf Z}_{1}^{N_S}$ and ${\bf Z'}_{1}^{N_S}$
be drawn from the source domain $\mathcal{Z}^{(S)}$, and ${\bf \overline{Z}}_{1}^{N_T}$ and ${\bf \overline{Z'}}_{1}^{N_T}$
be drawn from the target domain $\mathcal{Z}^{(T)}$.
Then, for any $\tau\in[0,1)$ and given an arbitrary $\xi>(1-\tau)D_\mathcal{F}(S,T)$, we have for any $N_SN_T\geq\frac{8(b-a)^2}{(\xi')^2}$,
\begin{align}\label{eq:sym1.C}
    \mathrm{Pr}\left\{\sup_{f\in\mathcal{F}}\big|\mathrm{E}^{(T)}f
    -\mathrm{E}_\tau f\big|>\xi\right\}
    \leq2\mathrm{Pr}\left\{\sup_{f\in\mathcal{F}}\big|\mathrm{E'}_\tau f-\mathrm{E}_\tau f\big|>\frac{\xi'}{2}\right\}
\end{align}
with $\xi'=\xi-(1-\tau)D_\mathcal{F}(S,T).$
\end{theorem}

This theorem shows that for any $\xi>(1-\tau)D_\mathcal{F}(S,T)$, the probability of the event:
$$\sup_{f\in\mathcal{F}}\big|\mathrm{E}^{(T)}f-\mathrm{E}_\tau f\big|>\xi$$
can be bounded by using the probability of the event:
 $$\sup_{f\in\mathcal{F}}\big|\mathrm{E'}_\tau f-\mathrm{E}_\tau f\big|>\frac{\xi'}{2}$$
 that is only determined by the samples drawn from the source domain $\mathcal{Z}^{(S)}$ and the target domain $\mathcal{Z}^{(T)}$, when $N_SN_T\geq \frac{8(b-a)^2}{(\xi')^2}$. Compared to the classical symmetrization result under the assumption of same distribution \citep[see][]{Bousquet04}, there is a discrepancy term $(1-\tau)D_\mathcal{F}(S,T)$. The two results will coincide when the source and the target domains match, {\it i.e.}, $D_\mathcal{F}(S,T)=0$. The following is the proof of Theorem \ref{thm:sym.C}.

{\it Proof of Theorem \ref{thm:sym.C}.}
Let $\widehat{f}$ be the function achieving the supremum: $$\sup_{f\in\mathcal{F}}|\mathrm{E}^{(T)}f-\mathrm{E}_\tau f|$$ with respect to ${\bf Z}_{1}^{N_S}$ and ${\bf \overline{Z}}_{1}^{N_T}$.
 According to \eqref{eq:error.C} and \eqref{eq:distance}, we arrive at
\begin{align}\label{eq:sym00.M}
 \big|\mathrm{E}^{(T)}\widehat{f}-\mathrm{E}_\tau\widehat{f}\big|
 =&\big|\tau\mathrm{E}^{(T)}\widehat{f}+(1-\tau)\mathrm{E}^{(T)}\widehat{f}
 -(1-\tau)\mathrm{E}^{(S)}\widehat{f}
 +(1-\tau)\mathrm{E}^{(S)}\widehat{f}-\mathrm{E}_\tau\widehat{f}\big|\nonumber\\
 =&\big|\tau(\mathrm{E}^{(T)}\widehat{f}-\mathrm{E}_{N_T}^{(T)}\widehat{f})
+(1-\tau)(\mathrm{E}^{(T)}\widehat{f}-\mathrm{E}^{(S)}\widehat{f})
+(1-\tau)(\mathrm{E}^{(S)}\widehat{f}-\mathrm{E}_{N_S}^{(S)}
\widehat{f})\big|\nonumber\\
\leq&(1-\tau)D_\mathcal{F}(S,T)+\big|\tau(\mathrm{E}^{(T)}\widehat{f}-
\mathrm{E}_{N_T}^{(T)}\widehat{f})
+(1-\tau)(\mathrm{E}^{(S)}\widehat{f}-\mathrm{E}_{N_S}^{(S)}\widehat{f})\big|,
\end{align}
and thus
\begin{align}\label{eq:sym01.C}
 &\mathrm{Pr}\left\{\big|\mathrm{E}^{(T)}\widehat{f}-\mathrm{E}_\tau
 \widehat{f}\big|>\xi\right\}\nonumber\\
 \leq& \mathrm{Pr}\left\{(1-\tau)D_\mathcal{F}(S,T)+\big|\tau(\mathrm{E}^{(T)}
 \widehat{f}-\mathrm{E}_{N_T}^{(T)}\widehat{f})
+(1-\tau)(\mathrm{E}^{(S)}
\widehat{f}-\mathrm{E}_{N_S}^{(S)}\widehat{f})\big|>\xi\right\},
\end{align}
where
\begin{align}\label{eq:empirical.C}
  \mathrm{E}_{N_T}^{(T)}\widehat{f}
:=\frac{1}{N_T}\sum_{n=1}^{N_T}\widehat{f}({\bf z}^{(T)}_n);\quad
\mathrm{E}_{N_S}^{(S)}\widehat{f}
:=\frac{1}{N_S}\sum_{n=1}^{N_S}\widehat{f}({\bf z}^{(S)}_n).
\end{align}

Let
\begin{equation}\label{eq:xi1.C}
    \xi'=\xi-(1-\tau)D_\mathcal{F}(S,T)
\end{equation}
and denote $\wedge$ as the conjunction of two events. According to the triangle inequality, we have
\begin{align*}
 &\left(\mathbf{1}_{\left\{|\tau(\mathrm{E}^{(T)}\widehat{f}
 -\mathrm{E}_{N_T}^{(T)}\widehat{f})
+(1-\tau)(\mathrm{E}^{(S)}
\widehat{f}-\mathrm{E}_{N_S}^{(S)}\widehat{f})|>\xi'\right\}}\right)
\left(\mathbf{1}_{\left\{|\tau(\mathrm{E}^{(T)}\widehat{f}
-\mathrm{E'}_{N_T}^{(T)}\widehat{f})
+(1-\tau)(\mathrm{E}^{(S)}
\widehat{f}-\mathrm{E'}_{N_S}^{(S)}\widehat{f})|<\frac{\xi'}{2}\right\}}\right)\nonumber\\
=&\mathbf{1}_{\left\{|\tau(\mathrm{E}^{(T)}\widehat{f}
-\mathrm{E}_{N_T}^{(T)}\widehat{f})
+(1-\tau)(\mathrm{E}^{(S)}
\widehat{f}-\mathrm{E}_{N_S}^{(S)}\widehat{f})|>\xi'\right\}
\,\wedge\,\left\{|\tau(\mathrm{E}^{(T)}\widehat{f}-\mathrm{E'}_{N_T}^{(T)}\widehat{f}
)
+(1-\tau)(\mathrm{E}^{(S)}
\widehat{f})-\mathrm{E'}_{N_S}^{(S)}\widehat{f}|<\frac{\xi'}{2}\right\}}\nonumber\\
\leq&\mathbf{1}_{\left\{|\tau(\mathrm{E'}_{N_T}^{(T)}\widehat{f}
-\mathrm{E}_{N_T}^{(T)}\widehat{f})
+(1-\tau)(\mathrm{E'}_{N_S}^{(S)}
\widehat{f}-\mathrm{E}_{N_S}^{(S)}\widehat{f})|>\frac{\xi'}{2}\right\}}\;.
\end{align*}
Then, taking the expectation with respect to ${\bf Z'}_{1}^{N_S}$ and ${\bf \overline{Z'}}_{1}^{N_T}$ gives
\begin{align}\label{eq:the:sym.pr2.C}
   &\left(\mathbf{1}_{\left\{|\tau(\mathrm{E}^{(T)}\widehat{f}
   -\mathrm{E}_{N_T}^{(T)}\widehat{f})
+(1-\tau)(\mathrm{E}^{(S)}
\widehat{f}-\mathrm{E}_{N_S}^{(S)}\widehat{f})|>\xi'\right\}}\right)\nonumber\\
&\times\mathrm{Pr'}\left\{\big|\tau(\mathrm{E}^{(T)}\widehat{f}
-\mathrm{E'}_{N_T}^{(T)}\widehat{f})
+(1-\tau)(\mathrm{E}^{(S)}
\widehat{f}-\mathrm{E'}_{N_S}^{(S)}\widehat{f})\big|<\frac{\xi'}{2}\right\}\nonumber\\
&\leq \mathrm{Pr'}\left\{\big|\tau(\mathrm{E'}_{N_T}^{(T)}\widehat{f}
-\mathrm{E}_{N_T}^{(T)}\widehat{f})
+(1-\tau)(\mathrm{E'}_{N_S}^{(S)}
\widehat{f}-\mathrm{E}_{N_S}^{(S)}\widehat{f})\big|>\frac{\xi'}{2}\right\}.
\end{align}

By Chebyshev's inequality, since ${\bf Z'}_{1}^{N_S}=\{{\bf z'}^{(S)}_n\}_{n=1}^{N_S}$ and ${\bf \overline{Z'}}_{1}^{N_T}=\{{\bf z'}^{(T)}_n\}_{n=1}^{N_T}$ are sets of i.i.d. samples drawn from the source domain $\mathcal{Z}^{(S)}$ and the target domain $\mathcal{Z}^{(T)}$ respectively, we have for any $\xi'>0$ and any $\tau\in[0,1)$,
\begin{align}\label{eq:the:sym.pr3.C}
   &\mathrm{Pr'}\left\{\big|\tau(\mathrm{E}^{(T)}\widehat{f}
   -\mathrm{E'}_{N_T}^{(T)}\widehat{f})
+(1-\tau)(\mathrm{E}^{(S)}
\widehat{f}-\mathrm{E'}_{N_S}^{(S)}
\widehat{f})\big|\geq\frac{\xi'}{2}\right\}\nonumber\\
\leq&\mathrm{Pr'}\left\{\frac{\tau}{N_T}\sum_{n=1}^{N_T}
|\mathrm{E}^{(T)}\widehat{f}-\widehat{f}({\bf z'}_n^{(T)})|+\frac{1-\tau}{N_S}\sum_{n=1}^{N_S}|\mathrm{E}^{(S)}
\widehat{f}-\widehat{f}({\bf z'}_n^{(S)})|\geq\frac{\xi'}{2}\right\}\nonumber\\
\leq& \frac{4\mathrm{E}\left\{\tau N_SN_T(\mathrm{E}^{(T)}\widehat{f}-\widehat{f}({\bf z'}^{(T)}))^2+(1-\tau) N_SN_T(\mathrm{E}^{(S)}\widehat{f}-\widehat{f}({\bf z'}^{(S)}))^2\right\}}{N_S^2N_T^2(\xi')^2}\nonumber\\
\leq& \frac{4\mathrm{E}\left\{\tau N_SN_T(b-a)^2+(1-\tau) N_SN_T(b-a)^2\right\}}{N_S^2N_T^2(\xi')^2}\nonumber\\
=&\frac{4(b-a)^2}{N_SN_T(\xi')^2},
\end{align}
where ${\bf z'}^{(S)}$ and ${\bf z'}^{(T)}$ stand for the ghost random variables taking values from the source domain $\mathcal{Z}^{(S)}$ and the target domain $\mathcal{Z}^{(T)}$, respectively.

Subsequently, according to \eqref{eq:the:sym.pr2.C} and \eqref{eq:the:sym.pr3.C}, we have for any $\xi'>0$,
\begin{align}\label{eq:the:sym.pr4.C}
   &\mathrm{Pr'}\left\{\big|\tau(\mathrm{E'}_{N_T}^{(T)}\widehat{f}
   -\mathrm{E}_{N_T}^{(T)}\widehat{f})
+(1-\tau)(\mathrm{E'}_{N_S}^{(S)}
\widehat{f}-\mathrm{E}_{N_S}^{(S)}\widehat{f})\big|>\frac{\xi'}{2}\right\}\nonumber\\
&\geq
  \left(\mathbf{1}_{\left\{\big|\tau(\mathrm{E}^{(T)}\widehat{f}
  -\mathrm{E}_{N_T}^{(T)}\widehat{f})
+(1-\tau)(\mathrm{E}^{(S)}
\widehat{f}-\mathrm{E}_{N_S}^{(S)}\widehat{f})\big|>\xi'\right\}}\right)
\left(1-\frac{4(b-a)^2}
{N_SN_T(\xi')^2}\right).
\end{align}

According to \eqref{eq:sym01.C}, \eqref{eq:xi1.C} and \eqref{eq:the:sym.pr4.C}, by letting
$$\frac{4(b-a)^2}{N_SN_T(\xi')^2}\leq \frac{1}{2},$$
and taking the expectation with respect to ${\bf Z}_{1}^{N_S}$ and ${\bf \overline{Z}}_{1}^{N_T}$,
we have for any $\xi'>0$,
\begin{align}\label{eq:sym001.C}
&\mathrm{Pr}\left\{|\mathrm{E}^{(T)}\widehat{f}
-\mathrm{E}_\tau\widehat{f}|>\xi\right\}\nonumber\\
 \leq& \mathrm{Pr}\left\{\big|\tau(\mathrm{E}^{(T)}\widehat{f}
 -\mathrm{E}_{N_T}^{(T)}\widehat{f})
+(1-\tau)(\mathrm{E}^{(S)}
\widehat{f}-\mathrm{E}_{N_S}^{(S)}\widehat{f})\big|>\xi'\right\}\nonumber\\
\leq&2\mathrm{Pr}\left\{\big|\tau(\mathrm{E'}_{N_T}^{(T)}\widehat{f}
-\mathrm{E}_{N_T}^{(T)}\widehat{f})
+(1-\tau)(\mathrm{E'}_{N_S}^{(S)}
\widehat{f}-\mathrm{E}_{N_S}^{(S)}\widehat{f})\big|>\frac{\xi'}{2}\right\}\nonumber\\
=&2\mathrm{Pr}\left\{\big|\mathrm{E'}_\tau \widehat{f}-\mathrm{E}_\tau\widehat{f}\big|>\frac{\xi'}{2}\right\}
\end{align}
with $\xi'=\xi-(1-\tau)D_{\mathcal{F}}(S,T)$.
This completes the proof.
\hfill$\blacksquare$

We are now ready to prove Theorem \ref{thm:crate.C}.

\subsection{Proof of Theorem \ref{thm:crate.C}}

{\it Proof of Theorem \ref{thm:crate.C}.} Consider $\{\epsilon_n\}_{n=1}^N$ as independent Rademacher random variables, {\it i.e.}, independent $\{\pm1\}$-valued random variables with equal probability of taking either value. Given $\{\epsilon_n\}_{n=1}^{N_S}$, $\{\epsilon_n\}_{n=1}^{N_T}$, ${\bf Z}_{1}^{2N_S}$ and ${\bf \overline{Z}}_{1}^{2N_T}$, denote
\begin{align}\label{eq:epsilon.C}
    \overrightarrow{\epsilon}_S:=&(\epsilon_1,\cdots,\epsilon_{N_S},
    -\epsilon_1,\cdots,-\epsilon_{N_S})
\in\{\pm 1\}^{2N_S};\nonumber\\
\overrightarrow{\epsilon}_T:=&(\epsilon_1,\cdots,\epsilon_{N_T},
-\epsilon_1,\cdots,-\epsilon_{N_T})\in\{\pm 1\}^{2N_T},
\end{align}
and for any $f\in\mathcal{F}$,
\begin{align}\label{eq:vecf.C}
  \overrightarrow{f}({\bf Z}_{1}^{2N_S}):=&\big(f({\bf z}'_1),\cdots,f({\bf z}'_{N_S}),f({\bf z}_1),\cdots,f({\bf z}_{N_S})\big)\in[a,b]^{2N_S};\nonumber\\
\overrightarrow{f}({\bf Z}_{1}^{2N_T}):=&\big(f({\bf z}'_1),\cdots,f({\bf z}'_{N_T}),f({\bf z}_1),\cdots,f({\bf z}_{N_T})\big)\in[a,b]^{2N_T}.
\end{align}
We also denote
\begin{align}\label{eq:ep&f.C}
{\bf Z}:=&\left\{{\bf \overline{Z}}_{1}^{2N_T},{\bf Z}_{1}^{2N_S}\right\}\in\big(\mathcal{Z}^{(T)}\big)^{2N_T}
\times\big(\mathcal{Z}^{(S)}\big)^{2N_S};\nonumber\\
 \overrightarrow{\epsilon}:=&\Big(  \underbrace{\overrightarrow{\epsilon}_T,\cdots,
 \overrightarrow{\epsilon}_T}_{N_S},
\underbrace{\overrightarrow{\epsilon}_S,\cdots,
\overrightarrow{\epsilon}_S}_{N_T}\Big)\in\{\pm1\}^{4N_SN_T};\nonumber\\
\overrightarrow{f}\big({\bf Z}\big):=&\Big(\underbrace{\overrightarrow{f}({\bf \overline{Z}}_{1}^{2N_T}),\cdots,\overrightarrow{f}({\bf \overline{Z}}_{1}^{2N_T})}_{N_S},\underbrace{\overrightarrow{f}({\bf Z}_{1}^{2N_S}),\cdots,\overrightarrow{f}({\bf Z}_{1}^{2N_S})}_{N_T}\Big)\in[a,b]^{4N_SN_T}.
\end{align}

According to \eqref{eq:short1.M}, \eqref{eq:xi1.C} and Theorem \ref{thm:sym.C},  for any $\tau\in[0,1)$ and given an arbitrary $\xi>(1-\tau)D_\mathcal{F}(S,T)$, we have for any $N_SN_T\geq\frac{8(b-a)^2}{{\xi'}^2}$ with $\xi'=\xi-(1-\tau)D_\mathcal{F}(S,T)$,
\begin{align}\label{eq:bas1.C}
    &\mathrm{Pr}\left\{\sup_{f\in\mathcal{F}}\big|\mathrm{E}^{(T)}f-\mathrm{E}_\tau f\big|>\xi\right\}\nonumber\\
\leq& 2\mathrm{Pr}\left\{\sup_{f\in\mathcal{F}}
\big|\mathrm{E'}_\tau f-\mathrm{E}_\tau f\big|>\frac{\xi'}{2}\right\}\qquad\mbox{(by Theorem \ref{thm:sym.C})}\nonumber\\
=&2\mathrm{Pr}\left\{\sup_{f\in\mathcal{F}}\Big|\frac{\tau}{N_T}
\sum_{n=1}^{N_T}\big(f({\bf z'}^{(T)}_n)-f({\bf z}^{(T)}_n)\big)+\frac{1-\tau}{N_S}\sum_{n=1}^{N_S}\big(f({\bf z'}^{(S)}_n)-f({\bf z}^{(S)}_n)\big)\Big|>\frac{\xi'}{2}\right\}\nonumber\\
=&2\mathrm{Pr}\left\{\sup_{f\in\mathcal{F}}\Big|\frac{\tau}{N_T}
\sum_{n=1}^{N_T}\epsilon_n\big(f({\bf z'}^{(T)}_n)-f({\bf z}^{(T)}_n)\big)+\frac{1-\tau}{N_S}\sum_{n=1}^{N_S}\epsilon_n\big(f({\bf z'}^{(S)}_n)-f({\bf z}^{(S)}_n)\big)\Big|>\frac{\xi'}{2}\right\}\nonumber\\
=&2\mathrm{Pr}\left\{\sup_{f\in\mathcal{F}}\Big|\frac{\tau}{2N_T}
\big\langle\overrightarrow{\epsilon}_T,
\overrightarrow{f}({\bf \overline{Z}}_{1}^{2N_T})\big\rangle+\frac{1-\tau}{2N_S}
\big\langle\overrightarrow{\epsilon}_S,
\overrightarrow{f}({\bf Z}_{1}^{2N_S})\big\rangle\Big|>\frac{\xi'}{4}\right\}.
\end{align}
Given a $\tau\in[0,1)$, fix a realization of ${\bf Z}$ and let $\Lambda$ be a $\xi'/8$-radius cover of $\mathcal{F}$ with respect to the $\ell_1^\tau({\bf Z})$ norm.
Since $\mathcal{F}$ is composed of bounded functions with the range $[a,b]$, we assume that the same holds for any $h\in\Lambda$. If $f_0$ is the function that achieves the following supremum $$\sup_{f\in\mathcal{F}}\Big|\frac{\tau}{2N_T}
\big\langle\overrightarrow{\epsilon}_T,
\overrightarrow{f}({\bf \overline{Z}}_{1}^{2N_T})\big\rangle+\frac{1-\tau}{2N_S}
\big\langle\overrightarrow{\epsilon}_S,
\overrightarrow{f}({\bf Z}_{1}^{2N_S})\big\rangle\Big|>\frac{\xi'}{4},$$
 there must be an $h_0\in\Lambda$ that satisfies that
\begin{align*}
    &\frac{\tau}{2N_T}\sum_{n=1}^{N_T}\left(|f_0({\bf z'}^{(T)}_n)-h_0({\bf z'}^{T}_n)|+|f_0({\bf z}^{(T)}_n)-h_0({\bf z}^{(T)}_n)|\right)\nonumber\\
&+\frac{1-\tau}{2N_S}\sum_{n=1}^{N_S}\left(|f_0({\bf z'}^{(S)}_n)-h_0({\bf z'}^{S}_n)|+|f_0({\bf z}^{(S)}_n)-h_0({\bf z}^{(S)}_n)|\right)< \frac{\xi'}{8},
\end{align*}
and meanwhile,
\begin{equation*}
    \Big|\frac{\tau}{2N_T}\big\langle\overrightarrow{\epsilon}_T,
\overrightarrow{h}_0({\bf \overline{Z}}_{1}^{2N_T})\big\rangle+\frac{1-\tau}{2N_S}
\big\langle\overrightarrow{\epsilon}_S,
\overrightarrow{h}_0({\bf Z}_{1}^{2N_S})\big\rangle\Big|>\frac{\xi'}{8}.
\end{equation*}
Therefore, for the realization of ${\bf Z}$, we arrive at
\begin{align}\label{eq:bas2.C}
&\mathrm{Pr}\left\{\sup_{f\in\mathcal{F}}\Big|\frac{\tau}{2N_T}
\big\langle\overrightarrow{\epsilon}_T,
\overrightarrow{f}({\bf \overline{Z}}_{1}^{2N_T})\big\rangle+\frac{1-\tau}{2N_S}
\big\langle\overrightarrow{\epsilon}_S,
\overrightarrow{f}({\bf Z}_{1}^{2N_S})\big\rangle\Big|>\frac{\xi'}{4}\right\}\nonumber\\
\leq&\mathrm{Pr}\left\{\sup_{h\in\Lambda}
\Big|\frac{\tau}{2N_T}\big\langle\overrightarrow{\epsilon}_T,
\overrightarrow{h}({\bf \overline{Z}}_{1}^{2N_T})\big\rangle
+\frac{1-\tau}{2N_S}\big\langle\overrightarrow{\epsilon}_S,
\overrightarrow{h}({\bf Z}_{1}^{2N_S})\big\rangle\Big|>\frac{\xi'}{8}\right\}.
\end{align}

Moreover, we denote the event
\begin{equation*}
A:=\left\{\mathrm{Pr}\left\{\sup_{h\in\Lambda}
\Big|\frac{\tau}{2N_T}\big\langle\overrightarrow{\epsilon}_T,
\overrightarrow{h}({\bf \overline{Z}}_{1}^{2N_T})\big\rangle+\frac{1-\tau}{2N_S}
\big\langle\overrightarrow{\epsilon}_S,
\overrightarrow{h}({\bf Z}_{1}^{2N_S})\big\rangle\Big|>\frac{\xi'}{8}\right\}\right\},
\end{equation*}
and let ${\bf 1}_A$ be the characteristic function of the event $A$. By Fubini's Theorem, we have
\begin{align}\label{eq:Fubini.C}
&\mathrm{Pr}\{A\}=\mathrm{E}\Big\{\mathrm{E}_{\overrightarrow{\epsilon}}\big\{{\bf 1}_A\big\}\big|\;{\bf Z}\Big\}\nonumber\\
=&\mathrm{E}\left\{\mathrm{Pr}\left\{\sup_{h\in\Lambda}
\Big|\frac{\tau}{2N_T}\big\langle\overrightarrow{\epsilon}_T,
\overrightarrow{h}({\bf \overline{Z}}_{1}^{2N_T})\big\rangle+\frac{1-\tau}{2N_S}
\big\langle\overrightarrow{\epsilon}_S,
\overrightarrow{h}({\bf Z}_{1}^{2N_S})\big\rangle\Big|>\frac{\xi'}{8}\right\}\big|\;{\bf Z}\right\}.
\end{align}

Fix a realization of ${\bf Z}$ again. According to \eqref{eq:L1_tau}, \eqref{eq:epsilon.C}, \eqref{eq:vecf.C} and Theorem \ref{thm:dineq.C}, for any $\tau\in[0,1)$ and given an arbitrary $\xi'>0$, we have for any $N_SN_T\geq \frac{8(b-a)^2}{(\xi')^2}$,
\begin{align}\label{eq:bas3.C}
&\mathrm{Pr}\left\{\sup_{h\in\Lambda}\Big|\frac{\tau}{2N_T}
\big\langle\overrightarrow{\epsilon}_T,
\overrightarrow{h}({\bf \overline{Z}}_{1}^{2N_T})\big\rangle+\frac{1-\tau}{2N_S}
\big\langle\overrightarrow{\epsilon}_S,
\overrightarrow{h}({\bf Z}_{1}^{2N_S})\big\rangle\Big|>\frac{\xi'}{8}\right\}\nonumber\\
\leq&|\Lambda|\max_{h\in\Lambda}\mathrm{Pr}\left\{
\Big|\frac{\tau}{2N_T}\big\langle\overrightarrow{\epsilon}_T,
\overrightarrow{h}({\bf \overline{Z}}_{1}^{2N_T})\big\rangle+\frac{1-\tau}{2N_S}
\big\langle\overrightarrow{\epsilon}_S,
\overrightarrow{h}({\bf Z}_{1}^{2N_S})\big\rangle\Big|>\frac{\xi'}{8}\right\}\nonumber\\
=&\mathcal{N}\left(\mathcal{F},\xi'/8,\ell^\tau_1({\bf Z})\right)\max_{h\in\Lambda}\mathrm{Pr}\left\{\big|\mathrm{E'}_\tau h-\mathrm{E}_\tau h\big|>
\frac{\xi'}{4}\right\}\nonumber\\
\leq& \mathcal{N}\left(\mathcal{F},\xi'/8,\ell^\tau_1({\bf Z})\right)\max_{h\in\Lambda}\mathrm{Pr}
\left\{|\widetilde{\mathrm{E}}h-\mathrm{E'}_\tau h|
+|\widetilde{\mathrm{E}}h-\mathrm{E}_\tau h|
>\frac{\xi'}{4}\right\}\nonumber\\
\leq&
2\mathcal{N}\left(\mathcal{F},\xi'/8,\ell^\tau_1({\bf Z})\right)\max_{h\in\Lambda}
\mathrm{Pr}\left\{\big|\widetilde{\mathrm{E}}h-\mathrm{E}_\tau h\big|>\frac{\xi'}{8}\right\}\nonumber\\
\leq&4\mathcal{N}\left(\mathcal{F},\xi'/8,\ell^\tau_1({\bf Z})\right)
\exp\left\{-\frac{N_SN_T\left(\xi-(1-\tau)D_{\mathcal{F}}
(S,T)\right)^2}{32(b-a)^2\left((1-\tau)^2N_T+\tau^2N_S\right)}\right\},
\end{align}
where $\widetilde{\mathrm{E}}h:=\tau\mathrm{E}^{(T)}h+(1-\tau)\mathrm{E}^{(S)}h$.

The combination of \eqref{eq:UEN2.C}, \eqref{eq:bas1.C}, \eqref{eq:bas2.C} and \eqref{eq:bas3.C} leads to the following result: for any $\tau\in[0,1)$ and given an arbitrary $\xi>(1-\tau)D_{\mathcal{F}}(S,T)$, we have for any $N_SN_T\geq \frac{8(b-a)^2}{(\xi')^2}$,
\begin{align}\label{eq:RB1_1.C}
    &\mathrm{Pr}\left\{\sup_{f\in\mathcal{F}}
    \big|\mathrm{E}^{(T)}f-\mathrm{E}_\tau f\big|>\xi\right\}\nonumber\\
        \leq&8\mathrm{E}\mathcal{N}\left(\mathcal{F},\xi'/8,\ell^\tau_1({\bf Z})\right)
        \exp\left\{-\frac{N_SN_T\left(\xi-(1-\tau)D_{\mathcal{F}}
        (S,T)\right)^2}{32(b-a)^2\left((1-\tau)^2N_T+\tau^2N_S\right)}\right\}\nonumber\\
        \leq&8\mathcal{N}_1^\tau(\mathcal{F},\xi'/8,2(N_S+N_T))
        \exp\left\{-\frac{N_SN_T\left(\xi-(1-\tau)D_{\mathcal{F}}
        (S,T)\right)^2}{32(b-a)^2\left((1-\tau)^2N_T+\tau^2N_S\right)}\right\}.
\end{align}

According to \eqref{eq:RB1_1.C}, letting
\begin{align*}
    \epsilon:=8\mathcal{N}_1^\tau(\mathcal{F},\xi'/8,2(N_S+N_T))
        \exp\left\{-\frac{N_SN_T\left(\xi-(1-\tau)D_{\mathcal{F}}
        (S,T)\right)^2}{32(b-a)^2\left((1-\tau)^2N_T+\tau^2N_S\right)}\right\},
\end{align*}
we have given an arbitrary $\xi>(1-\tau)D_{\mathcal{F}}(S,T)$ and for any $N_SN_T\geq\frac{8(b-a)^2}{(\xi')^2}$, with probability at least $1-\epsilon$,
\begin{align*}
    \sup_{f\in\mathcal{F}}
\big|\mathrm{E}_\tau f-\mathrm{E}^{(T)}f\big|
\leq&(1-\tau)D_{\mathcal{F}}(S,T)+ \left(\frac{\ln\mathcal{N}_1^\tau(\mathcal{F},\xi'/8,2(N_S+N_T))-\ln(\epsilon/8)}
{\frac{N_SN_T}{32(b-a)^2\left((1-\tau)^2N_T+\tau^2N_S\right)}}\right)^{\frac{1}{2}}.
\end{align*}
This completes the proof. \hfill$\blacksquare$


\section{Proofs of Theorems \ref{thm:RB.Rade} $\&$ \ref{thm:RB.Rade.C}}\label{app:Proof3}

In this appendix, we will prove Theorem \ref{thm:RB.Rade} and Theorem \ref{thm:RB.Rade.C}. In order to achieve the proofs, we need to generalize the classical McDiarmid's inequality \citep[see][Theorem 6]{Bousquet04} to a more general setting where independent random variables can independently take values from different domains.

\subsection{Generalized McDiarmid's inequality}

The following is the classical McDiarmid's inequality that is one of the most frequently used deviation inequalities in statistical learning theory and has been widely used to obtain generalization bounds based on the Rademacher complexity under the assumption of same distribution \citep[see][Theorem 6]{Bousquet04}.

\begin{theorem}[McDiamid's Inequality]\label{thm:Mcdiarmid0}
Let ${\bf z}_{1},\cdots,{\bf z}_{N}$ be $N$ independent random variables taking value from the domain $\mathcal{Z}$. Assume that the function $H:\mathcal{Z}\rightarrow\mathbb{R}$  satisfies the condition of bounded difference:
for all $1\leq n\leq N$,
\begin{align}\label{eq:dif.cond}
&\sup_{{\bf z}_{1},\cdots,{\bf z}_{N},{\bf z'}_n}\Big|H\big({\bf z}_{1},\cdots,{\bf z}_n,\cdots,{\bf z}_{N}\big)
-H\big({\bf z}_{1},\cdots,{\bf z'}_n,\cdots,{\bf z}_{N}\big)\Big|\leq c_n.
\end{align}
Then, for any $\xi>0$
\begin{equation*}
\mathrm{Pr}\left\{H\big({\bf z}_{1},\cdots,{\bf z}_n,\cdots,{\bf z}_{N}\big)-\mathrm{E}\Big\{H\big({\bf z}_{1},\cdots,{\bf z}_n,\cdots,{\bf z}_{N}\big)\Big\}\geq\xi\right\}
\leq\exp\left\{-2\xi^2/\sum_{n=1}^{N}c_n^2\right\}.
\end{equation*}
\end{theorem}
As shown in Theorem \ref{thm:Mcdiarmid0}, the classical McDiarmid's inequality is valid under the condition that random variables ${\bf z}_{1},\cdots,{\bf z}_{N}$ are independent and drawn from the same domain. Next, we generalize this inequality to a more general setting, where the independent random variables can take values from different domains.

\begin{theorem}\label{thm:Mcdiarmid}
Given independent domains $\mathcal{Z}^{(S_k)}$ ($1\leq k\leq K$), for any $1\leq k\leq K$, let ${\bf Z}_{1}^{N_k}:=\{{\bf z}_n^{(S_k)}\}_{n=1}^{N_k}$ be $N_k$ independent random variables taking values from the domain $\mathcal{Z}^{(S_k)}$. Assume that the function $H:\big(\mathcal{Z}^{(S_1)}\big)^{N_1}\times\cdots\times\big(\mathcal{Z}^{(S_K)}\big)^{N_K}\rightarrow\mathbb{R}$  satisfies the condition of bounded difference:
for all $1\leq k\leq K$ and $1\leq n\leq N_k$,
\begin{align}\label{eq:dif.cond}
&\sup_{{\bf Z}_{1}^{N_1},\cdots,{\bf Z}_{1}^{N_K},{\bf z'}^{(S_k)}_n}\Big|H\big({\bf Z}_{1}^{N_1},\cdots,{\bf Z}_{1}^{N_{k-1}},{\bf z}_1^{(S_k)},\cdots,{\bf z}_n^{(S_k)},\cdots,{\bf z}_{N_k}^{(S_k)},{\bf Z}_{1}^{N_{k+1}},\cdots,{\bf Z}_{1}^{N_K}\big)\nonumber\\
-&H\big({\bf Z}_{1}^{N_1},\cdots,{\bf Z}_{1}^{N_{k-1}},{\bf z}_1^{(S_k)},\cdots,{\bf z'}_n^{(S_k)},\cdots,{\bf z}_{N_k}^{(S_k)},{\bf Z}_{1}^{N_{k+1}},\cdots,{\bf Z}_{1}^{N_K}\big)\Big|\leq c^{(k)}_n.
\end{align}
Then, for any $\xi>0$
\begin{equation*}
\mathrm{Pr}\left\{H\big({\bf Z}_{1}^{N_1},\cdots,{\bf Z}_{1}^{N_K}\big)-\mathrm{E}\Big\{H\big({\bf Z}_{1}^{N_1},\cdots,{\bf Z}_{1}^{N_K}\big)\Big\}\geq\xi\right\}
\leq\exp\left\{-2\xi^2/\sum_{k=1}^{K}\sum_{n=1}^{N_k}(c^{(k)}_n)^2\right\}.
\end{equation*}
\end{theorem}

{\bf Proof.} Define a random variable
\begin{equation}\label{eq:Tn1}
    T^{(k)}_{n}:=\mathrm{E}\left\{H(\{{\bf Z}_{1}^{N_k}\}_{k=1}^K)|{\bf Z}_{1}^{N_1},{\bf Z}_{1}^{N_2},\cdots,{\bf Z}_{1}^{N_{k-1}},{\bf Z}_1^{n}\right\},\;\; 1\leq k\leq K,\;0\leq n\leq N_k,
\end{equation}
where
$${\bf Z}_{1}^{n}=\{{\bf z}^{(k)}_1,{\bf z}^{(k)}_2,\cdots,{\bf z}^{(k)}_{n}\}\subseteq{\bf Z}_{1}^{N_k},  \mbox{  and   }   {\bf Z}_{1}^0=\varnothing.$$
It is clear that
$$T_0^{(1)}=\mathrm{E}\big\{H(\{{\bf Z}_{1}^{N_k}\}_{k=1}^K)\big\} \mbox{  and }  T^{(K)}_{N_K}=H(\{{\bf Z}_{1}^{N_k}\}_{k=1}^K),$$
and thus
\begin{equation}\label{eq:Tn2}
 H(\{{\bf Z}_{1}^{N_k}\}_{k=1}^K)-\mathrm{E}\big\{H(\{{\bf Z}_{1}^{N_k}\}_{k=1}^K)\big\}
 =T^{(K)}_{N_K}-T_0^{(1)}
 =\sum_{k=1}^K\sum_{n=1}^{N_k}(T^{(k)}_{n}-T^{(k)}_{n-1}).
\end{equation}

Denote for any $1\leq k\leq K$ and $1\leq n\leq N_k$,
\begin{align}\label{eq:UL1}
 U_n^{(k)}=&\sup_{\mu}\left\{T^{(k)}_{n}\big|_{{\bf z}^{(k)}_n=\mu}-T^{(k)}_{n-1}\right\};\nonumber\\
 L_n^{(k)}=&\inf_{\nu}\left\{T^{(k)}_{n}\big|_{{\bf z}^{(k)}_n=\nu}-T^{(k)}_{n-1}\right\}.\nonumber
\end{align}
It follows from the definition of \eqref{eq:Tn1} that $L_n^{(k)}\leq(T^{(k)}_{n}-T^{(k)}_{n-1})\leq U_n^{(k)}$ and thus results in
\begin{equation}\label{eq:UL2}
T^{(k)}_{n}-T^{(k)}_{n-1}\leq U_n^{(k)}-L_n^{(k)}=\sup_{\mu,\nu}\left\{T^{(k)}_{n}\big|_{{\bf z}^{(k)}_n=\mu}-T^{(k)}_{n}\big|_{{\bf z}^{(k)}_n=\nu}\right\}\leq c^{(k)}_n.
\end{equation}
Moreover, by the law of iterated expectation, we also have for any $1\leq k\leq K$ and $1\leq n\leq N_k$
\begin{equation}\label{eq:UL3}
\mathrm{E}\left\{T^{(k)}_{n}-T^{(k)}_{n-1}|{\bf Z}_{1}^{N_1},{\bf Z}_{1}^{N_2},\cdots,{\bf Z}_{1}^{N_{k-1}},{\bf Z}_1^{n-1}\right\}=0.
\end{equation}
According to Hoeffding inequality \citep[see][]{Hoeffding63}, given an $\alpha>0$, the condition \eqref{eq:dif.cond} leads to for any $1\leq k\leq K$ and $1\leq n\leq N_k$,
\begin{equation}\label{eq:Tn3}
    \mathrm{E}\left\{\mathrm{e}^{\alpha(T^{(k)}_{n}-T^{(k)}_{n-1})}|{\bf Z}_{1}^{N_1},{\bf Z}_{1}^{N_2},\cdots,{\bf Z}_{1}^{N_{k-1}},{\bf Z}_1^{n-1}\right\}\leq\mathrm{e}^{\alpha^2(c^{(k)}_n)^2/8}.
\end{equation}

Subsequently, according to Markov's inequality, \eqref{eq:Tn2}, \eqref{eq:UL2}, \eqref{eq:UL3} and \eqref{eq:Tn3},  we have for any $\alpha>0$,
\begin{align}\label{eq:Main.Mc}
&\mathrm{Pr}\left\{H\big({\bf Z}_{1}^{N_1},\cdots,{\bf Z}_{1}^{N_K}\big)-\mathrm{E}\Big\{H\big({\bf Z}_{1}^{N_1},\cdots,{\bf Z}_{1}^{N_K}\big)\Big\}\geq\xi\right\}\nonumber\\
\leq&\mathrm{e}^{-\alpha\xi}\mathrm{E}\Big\{\mathrm{e}^{\alpha\big(H\big({\bf Z}_{1}^{N_1},\cdots,{\bf Z}_{1}^{N_K}\big)-\mathrm{E}\big\{H\big({\bf Z}_{1}^{N_1},\cdots,{\bf Z}_{1}^{N_K}\big)\big\}\big)}\Big\}\nonumber\\
=&\mathrm{e}^{-\alpha\xi}\mathrm{E}
\Big\{\mathrm{e}^{\alpha\big(\sum_{k=1}^K\sum_{n=1}^{N_k}(T^{(k)}_{n}
-T^{(k)}_{n-1})\big)}\Big\}\nonumber\\
=&\mathrm{e}^{-\alpha\xi}\mathrm{E}\left\{\mathrm{E}\left\{\mathrm{e}^{\alpha
\big(\sum_{k=1}^K\sum_{n=1}^{N_k}(T^{(k)}_{n}
-T^{(k)}_{n-1})\big)}\right\}|{\bf Z}_{1}^{N_1},\cdots,{\bf Z}_{1}^{N_{k-1}},{\bf Z}_1^{N_K-1} \right\}\nonumber\\
=&\mathrm{e}^{-\alpha\xi}\mathrm{E}\Big\{\mathrm{e}^{\alpha\big(\sum_{k=1}^{K-1}
\sum_{n=1}^{N_k}(T^{(k)}_{n}
-T^{(k)}_{n-1})+\sum_{n=1}^{N_K-1}(T^{(K)}_{n}
-T^{(K)}_{n-1})\big)}\Big\{\mathrm{e}^{\alpha\big(T^{(K)}_{N_K}
-T^{(K)}_{N_K-1}\big)}|{\bf Z}_{1}^{N_1},\cdots,{\bf Z}_{1}^{N_{K-1}},{\bf Z}_1^{N_K-1}\Big\} \Big\}\nonumber\\
\leq&\mathrm{e}^{-\alpha\xi}\mathrm{E}\Big\{\mathrm{e}^{\alpha\big(\sum_{k=1}^{K-1}
\sum_{n=1}^{N_k}(T^{(k)}_{n}
-T^{(k)}_{n-1})+\sum_{n=1}^{N_K-1}(T^{(K)}_{n}
-T^{(K)}_{n-1})\big)}\Big\}\mathrm{e}^{\alpha^2(c^{(K)}_{N_K})^2/8} \nonumber\\
\leq & \mathrm{e}^{-\alpha\xi}\prod_{k=1}^K\prod_{n=1}^{N_k}
\exp\left\{\frac{\alpha^2(c^{(k)}_n)^2}{8}\right\}\nonumber\\
=&\exp\left\{-\alpha\xi+\alpha^2\sum_{k=1}^K\sum_{n=1}^{N_k}
\frac{(c^{(k)}_n)^2}{8}\right\}.\nonumber
\end{align}
The above bound is minimized by setting $$\alpha^*=\frac{4\xi}{\sum_{k=1}^K\sum_{n=1}^{N_K}(c^{(k)}_n)^2},$$
and its minimum value is
$$\exp\left\{-2\xi^2/\sum_{k=1}^{K}\sum_{n=1}^{N_k}(c^{(k)}_n)^2\right\}.$$
This completes the proof. \hfill$\blacksquare$

\subsection{Proof of Theorem \ref{thm:RB.Rade}}

By using Theorem \ref{thm:Mcdiarmid}, we prove Theorem \ref{thm:RB.Rade} as follows:

{\bf Proof of Theorem \ref{thm:RB.Rade}} Assume that $\mathcal{F}$ is a function class $\mathcal{F}$ consisting of bounded functions with the range $[a,b]$. Let sample sets $\{{\bf Z}_{1}^{N_k}\}_{k=1}^K:=\{\{{\bf z}_{n}^{(k)}\}_{n=1}^{N_k}\}_{k=1}^K$ be drawn from multiple sources $\mathcal{Z}^{(S_k)}$ ($1\leq k\leq K$), respectively. Given a choice of ${\bf w}\in[0,1]^{K}$ with $\sum_{k=1}^Kw_k=1$, denote
\begin{equation}\label{eq:Proof.Rad1}
H\big({\bf Z}_{1}^{N_1},\cdots,{\bf Z}_{1}^{N_K}\big):=\sup_{f\in\mathcal{F}}\big|\mathrm{E}_{{\bf w}}^{(S)}f-\mathrm{E}^{(T)}f\big|.
\end{equation}
By \eqref{eq:emrisk.L}, we have
\begin{equation}\label{eq:Proof.Rad2}
H\big({\bf Z}_{1}^{N_1},\cdots,{\bf Z}_{1}^{N_K}\big)
=\sup_{f\in\mathcal{F}}\Big|\sum_{k=1}^Kw_k\big(\mathrm{E}_{N_k}^{(S)}f-\mathrm{E}^{(T)}f\big)\Big|,
\end{equation}
where $\mathrm{E}_{N_k}^{(S)}f=\frac{1}{N_k}\sum_{n=1}^{N_k}f({\bf z}_n^{(k)})$. Therefore, it is clear that such $H\big({\bf Z}_{1}^{N_1},\cdots,{\bf Z}_{1}^{N_K}\big)$ satisfies the condition of bounded difference with $$c_n^{(k)}=\frac{(b-a)w_k}{N_k}$$ for all $1\leq k\leq K$ and $1\leq n\leq N_k$. Thus, according to Theorem \ref{thm:Mcdiarmid}, we have for any $\xi>0$,
\begin{equation*}
\mathrm{Pr}\Big\{H\big({\bf Z}_{1}^{N_k},\cdots,{\bf Z}_{1}^{N_k}\big)-\mathrm{E}^{(S)}\big\{H\big({\bf Z}_{1}^{N_k},\cdots,{\bf Z}_{1}^{N_k}\big)\big\}\geq\xi\Big\}\leq\exp\left\{\frac{-2\xi^2}
{\sum_{k=1}^K\frac{(b-a)^2w_k^2}{N_k}}\right\},
\end{equation*}
which can be equivalently rewritten as with probability at least $1-\epsilon$,
\begin{align}\label{eq:Proof.Rad3}
H\big({\bf Z}_{1}^{N_k},\cdots,{\bf Z}_{1}^{N_k}\big)\leq&\mathrm{E}^{(S)}\big\{H\big({\bf Z}_{1}^{N_k},\cdots,{\bf Z}_{1}^{N_k}\big)\big\}+
\sqrt{\sum_{k=1}^K\frac{(b-a)^2w_k^2\ln(1/\epsilon)}{2N_k}}\nonumber\\
=&\mathrm{E}^{(S)}
   \left\{\sup_{f\in\mathcal{F}}\Big|\sum_{k=1}^Kw_k\big(\mathrm{E}_{N_k}^{(S)}f
   -\mathrm{E}^{(T)}f\big)\Big|\right\}+
\sqrt{\sum_{k=1}^K\frac{(b-a)^2w_k^2\ln(1/\epsilon)}{2N_k}}\nonumber\\
=&\mathrm{E}^{(S)}
   \left\{\sup_{f\in\mathcal{F}}\Big|\sum_{k=1}^Kw_k\big(\mathrm{E}_{N_k}^{(S)}f
   -\mathrm{E}^{(S_k)}f+\mathrm{E}^{(S_k)}f-\mathrm{E}^{(T)}f\big)\Big|\right\}\nonumber\\
   &+
\sqrt{\sum_{k=1}^K\frac{(b-a)^2w_k^2\ln(1/\epsilon)}{2N_k}}\nonumber\\
\leq&\mathrm{E}^{(S)}
   \left\{\sup_{f\in\mathcal{F}}\Big|\sum_{k=1}^Kw_k\big(\mathrm{E}_{N_k}^{(S)}f
   -\mathrm{E}^{(S_k)}f\big)\Big|\right\}
   +
   \sum_{k=1}^Kw_k\sup_{f\in\mathcal{F}}\big|\mathrm{E}^{(S_k)}f-\mathrm{E}^{(T)}f\big|\nonumber\\
   &+
\sqrt{\sum_{k=1}^K\frac{(b-a)^2w_k^2\ln(1/\epsilon)}{2N_k}}\nonumber\\
\leq&\mathrm{E}^{(S)}
   \left\{\sup_{f\in\mathcal{F}}\Big|\sum_{k=1}^Kw_k\big(\mathrm{E}_{N_k}^{(S)}f
   -\mathrm{E}^{(S_k)}f\big)\Big|\right\}
   +
   D^{({\bf w})}_{\mathcal{F}}(S,T)\quad\mbox{[see \eqref{eq:Dist.M}]
   }\nonumber\\
   &+
\sqrt{\sum_{k=1}^K\frac{(b-a)^2w_k^2\ln(1/\epsilon)}{2N_k}}.
\end{align}

Next, according to \eqref{eq:ExRade} and \eqref{eq:Proof.Rad2}, we have
\begin{align}\label{eq:Proof.Rad4}
   &\mathrm{E}^{(S)}\sup_{f\in\mathcal{F}}\Big|\sum_{k=1}^Kw_k\big(\mathrm{E}_{N_k}^{(S)}f
   -\mathrm{E}^{(S_k)}f\big)\Big|\nonumber\\
   =&\mathrm{E}^{(S)}\sup_{f\in\mathcal{F}}\Big|\sum_{k=1}^Kw_k\big(\mathrm{E}_{N_k}^{(S)}f-\mathrm{E'}^{(S_k)}
   \{\mathrm{E'}_{N_k}^{(S)}f\}\big)\Big|\nonumber\\
   \leq&\mathrm{E}^{(S)}\mathrm{E'}^{(S)}\sup_{f\in\mathcal{F}}
   \Big|\sum_{k=1}^Kw_k\big(\mathrm{E}_{N_k}^{(S)}f-\mathrm{E'}_{N_k}^{(S)}f\big)\Big|\nonumber\\
   =&\mathrm{E}^{(S)}\mathrm{E'}^{(S)}\sup_{f\in\mathcal{F}}
   \Big|\sum_{k=1}^Kw_k\frac{1}{N_k}\sum_{n=1}^{N_k}\big(f({\bf z}_n^{(k)})-f({\bf z'}_n^{(k)})\big)\Big|\nonumber\\
   \leq&\mathrm{E}^{(S)}\mathrm{E'}^{(S)}\mathrm{E}_{\sigma}\sup_{f\in\mathcal{F}}
   \Big|\sum_{k=1}^Kw_k\frac{1}{N_k}\sum_{n=1}^{N_k}\sigma_{n}^{(k)}\big(f({\bf z}_n^{(k)})-f({\bf z'}_n^{(k)})\big)\Big|\nonumber\\
   \leq&2\mathrm{E}^{(S)}\mathrm{E}_{\sigma}\sup_{f\in\mathcal{F}}
   \Big|\sum_{k=1}^Kw_k\frac{1}{N_k}\sum_{n=1}^{N_k}\sigma_{n}^{(k)}f({\bf z}_n^{(k)})\Big|\nonumber\\
   \leq&2\mathrm{E}^{(S)}\mathrm{E}_{\sigma}\sum_{k=1}^Kw_k\frac{1}{N_k}\sup_{f\in\mathcal{F}}
   \Big|\sum_{n=1}^{N_k}\sigma_{n}^{(k)}f({\bf z}_n^{(k)})\Big|\nonumber\\
   =&2\sum_{k=1}^Kw_k\mathcal{R}^{(k)}(\mathcal{F}).
\end{align}
By combining \eqref{eq:Proof.Rad1}, \eqref{eq:Proof.Rad3} and \eqref{eq:Proof.Rad4}, we obtain with probability at least $1-\epsilon$
\begin{equation*}
\sup_{f\in\mathcal{F}}\big|\mathrm{E}_{{\bf w}}^{(S)}f-\mathrm{E}^{(T)}f\big|\leq D^{({\bf w})}_{\mathcal{F}}(S,T)+2\sum_{k=1}^Kw_k\mathcal{R}^{(k)}(\mathcal{F})+
\sqrt{\sum_{k=1}^K\frac{(b-a)^2w_k^2\ln(1/\epsilon)}{2N_k}}.
\end{equation*}
This completes the proof. \hfill$\blacksquare$


\subsection{Proof of Theorem \ref{thm:RB.Rade.C}}

In order to prove Theorem \ref{thm:RB.Rade.C}, we also need the following result \citep[see][Theorem 5]{Bousquet04}:

\begin{theorem}\label{thm:b2}
Let $\mathcal{F}\subseteq[a,b]^{\mathcal{Z}}$. For any $\epsilon>0$, with probability at least $1-\epsilon$, there holds that for any $f\in\mathcal{F}$,
\begin{align}\label{eq:bound.T2}
\mathrm{E}f\leq&\mathrm{E}_Nf+2\mathcal{R}(\mathcal{F})+\sqrt{\frac{(b-a)\ln(1/\epsilon)}{2N}}\nonumber\\
\leq& \mathrm{E}_Nf+2\mathcal{R}_N(\mathcal{F})+3\sqrt{\frac{(b-a)\ln(2/\epsilon)}{2N}}.
\end{align}
\end{theorem}

Again, we prove Theorem \ref{thm:RB.Rade.C} by using Theorems \ref{thm:Mcdiarmid} and \ref{thm:b2}.

{\bf Proof of Theorem \ref{thm:RB.Rade.C}}
 We only consider the result of Theorem \ref{thm:Mcdiarmid} in a special setting of $K=2$, $N_1=N_T$, $N_2=N_S$, $w_1=\tau$ and $w_2=1-\tau$.
Given a choice of $\tau\in[0,1)$, denote
\begin{equation}\label{eq:Proof.Rad1.C}
H\big({\bf Z}_{1}^{N_S},\overline{{\bf Z}}_{1}^{N_T}\big):=\sup_{f\in\mathcal{F}}\big|\mathrm{E}_{\tau}f-\mathrm{E}^{(T)}f\big|.
\end{equation}
By \eqref{eq:error.C}, we have
\begin{equation}\label{eq:Proof.Rad2.C}
H\big({\bf Z}_{1}^{N_S},\overline{{\bf Z}}_{1}^{N_T}\big)
=\sup_{f\in\mathcal{F}}\Big|\tau\mathrm{E}^{(T)}_{N_T}f+(1-\tau)\mathrm{E}^{(S)}_{N_S}f-\mathrm{E}^{(T)}f\Big|,
\end{equation}

According to Theorem \ref{thm:Mcdiarmid}, we have for any
$\xi>0$,
\begin{equation*}
\mathrm{Pr}\Big\{H\big({\bf Z}_{1}^{N_S},\overline{{\bf Z}}_{1}^{N_T}\big)-\mathrm{E}^{(S)}\big\{H\big({\bf Z}_{1}^{N_S},\overline{{\bf Z}}_{1}^{N_T}\big)\big\}\geq\xi\Big\}
\leq\exp\left\{\frac{-2\xi^2}{(b-a)^2\left(\frac{\tau^2}{N_T}+\frac{(1-\tau)^2}{N_S}\right)}\right\},
\end{equation*}
which can be equivalently rewritten as with probability at least $1-(\epsilon/2)$,
\begin{align*}
&H\big({\bf Z}_{1}^{N_S},\overline{{\bf Z}}_{1}^{N_T}\big)\nonumber\\
\leq&\mathrm{E}^{(S)}\big\{H\big({\bf Z}_{1}^{N_S},\overline{{\bf Z}}_{1}^{N_T}\big)\big\}+
\sqrt{\frac{(b-a)^2\ln(2/\epsilon)}{2}\left(\frac{\tau^2}{N_T}+\frac{(1-\tau)^2}{N_S}\right)}\nonumber\\
=&\mathrm{E}^{(S)}
   \left\{\sup_{f\in\mathcal{F}}\Big|\tau\mathrm{E}^{(T)}_{N_T}f+(1-\tau)\mathrm{E}^{(S)}_{N_S}f-\mathrm{E}^{(T)}f\Big|\right\}+
\sqrt{\frac{(b-a)^2\ln(2/\epsilon)}{2}\left(\frac{\tau^2}{N_T}+\frac{(1-\tau)^2}{N_S}\right)}\nonumber\\
=&\mathrm{E}^{(S)}
   \left\{\sup_{f\in\mathcal{F}}\Big|\tau\big(\mathrm{E}^{(T)}_{N_T}f-\mathrm{E}^{(T)}f)+(1-\tau)\big(\mathrm{E}^{(S)}_{N_S}f-\mathrm{E}^{(T)}f\big)\Big|\right\}\nonumber\\
   &+
\sqrt{\frac{(b-a)^2\ln(2/\epsilon)}{2}\left(\frac{\tau^2}{N_T}+\frac{(1-\tau)^2}{N_S}\right)}\nonumber\\
\leq&\tau\sup_{f\in\mathcal{F}}\Big|\mathrm{E}^{(T)}_{N_T}f-\mathrm{E}^{(T)}f\Big|+(1-\tau)\mathrm{E}^{(S)}
   \left\{\sup_{f\in\mathcal{F}}\Big|\mathrm{E}^{(S)}_{N_S}f-\mathrm{E}^{(T)}f\Big|\right\}\nonumber\\
   &+
\sqrt{\frac{(b-a)^2\ln(2/\epsilon)}{2}\left(\frac{\tau^2}{N_T}+\frac{(1-\tau)^2}{N_S}\right)}\nonumber\\
=&\tau\sup_{f\in\mathcal{F}}\Big|\mathrm{E}^{(T)}_{N_T}f-\mathrm{E}^{(T)}f\Big|+(1-\tau)\mathrm{E}^{(S)}
   \left\{\sup_{f\in\mathcal{F}}\Big|\mathrm{E}^{(S)}_{N_S}f-\mathrm{E}^{(S)}f+\mathrm{E}^{(S)}f-\mathrm{E}^{(T)}f\Big|\right\}\nonumber\\
   &+
\sqrt{\frac{(b-a)^2\ln(2/\epsilon)}{2}\left(\frac{\tau^2}{N_T}+\frac{(1-\tau)^2}{N_S}\right)}\nonumber\\
\leq&\tau\sup_{f\in\mathcal{F}}\Big|\mathrm{E}^{(T)}_{N_T}f-\mathrm{E}^{(T)}f\Big|+(1-\tau)\mathrm{E}^{(S)}
   \left\{\sup_{f\in\mathcal{F}}\Big|\mathrm{E}^{(S)}_{N_S}f-\mathrm{E}^{(S)}f\Big|\right\}\nonumber\\
   &+(1-\tau)\sup_{f\in\mathcal{F}}\Big|\mathrm{E}^{(S)}f-\mathrm{E}^{(T)}f\Big|+
\sqrt{\frac{(b-a)^2\ln(2/\epsilon)}{2}\left(\frac{\tau^2}{N_T}+\frac{(1-\tau)^2}{N_S}\right)}\nonumber
\end{align*}
\begin{align}\label{eq:Proof.Rad3.C}
=&\tau\sup_{f\in\mathcal{F}}\Big|\mathrm{E}^{(T)}_{N_T}f-\mathrm{E}^{(T)}f\Big|+(1-\tau)\mathrm{E}^{(S)}
   \left\{\sup_{f\in\mathcal{F}}\Big|\mathrm{E}^{(S)}_{N_S}f-\mathrm{E}^{(S)}f\Big|\right\}+(1-\tau)D_{\mathcal{F}}(S,T)\nonumber\\
   &+
\sqrt{\frac{(b-a)^2\ln(2/\epsilon)}{2}\left(\frac{\tau^2}{N_T}+\frac{(1-\tau)^2}{N_S}\right)}.
\end{align}

According to Theorem \ref{thm:b2}, for any $\epsilon>0$, we have with at least $1-(\epsilon/2)$,
\begin{equation}\label{eq:Proof.Rad3.5.C}
\sup_{f\in\mathcal{F}}\Big|\mathrm{E}^{(T)}_{N_T}f
-\mathrm{E}^{(T)}f\Big|\leq2\mathcal{R}^{(T)}_{N_T}(\mathcal{F})
+3\sqrt{\frac{(b-a)\ln(4/\epsilon)}{2N_T}}.
\end{equation}

Next, according to \eqref{eq:ExRade}, we have
\begin{align}\label{eq:Proof.Rad4.C}
   &\mathrm{E}^{(S)}
   \left\{\sup_{f\in\mathcal{F}}\Big|\mathrm{E}^{(S)}_{N_S}f-\mathrm{E}^{(S)}f\Big|\right\}\nonumber\\
   =&\mathrm{E}^{(S)}\sup_{f\in\mathcal{F}}\Big|\mathrm{E}_{N_S}^{(S)}f-\mathrm{E'}^{(S)}
   \{\mathrm{E'}_{N_S}^{(S)}f\}\Big|\nonumber\\
   \leq&\mathrm{E}^{(S)}\mathrm{E'}^{(S)}\sup_{f\in\mathcal{F}}
   \Big|\mathrm{E}_{N_k}^{(S)}f-\mathrm{E'}_{N_k}^{(S)}f\Big|\nonumber\\
   =&\mathrm{E}^{(S)}\mathrm{E'}^{(S)}\sup_{f\in\mathcal{F}}
   \Big|\frac{1}{N_S}\sum_{n=1}^{N_S}\big(f({\bf z}_n^{(S)})-f({\bf z'}_n^{(S)})\big)\Big|\nonumber\\
   \leq&\mathrm{E}^{(S)}\mathrm{E'}^{(S)}\mathrm{E}_{\sigma}\sup_{f\in\mathcal{F}}
   \Big|\frac{1}{N_S}\sum_{n=1}^{N_S}\sigma_{n}\big(f({\bf z}_n^{(S)})-f({\bf z'}_n^{(S)})\big)\Big|\nonumber\\
   \leq&2\mathrm{E}^{(S)}\mathrm{E}_{\sigma}\sup_{f\in\mathcal{F}}
   \Big|\frac{1}{N_S}\sum_{n=1}^{N_S}\sigma_{n}f({\bf z}_n^{(S)})\Big|\nonumber\\
   =&2\mathcal{R}^{(S)}(\mathcal{F}).
\end{align}
By combining \eqref{eq:Proof.Rad1.C}, \eqref{eq:Proof.Rad3.C}, \eqref{eq:Proof.Rad3.5.C} and \eqref{eq:Proof.Rad4.C}, we obtain with probability at least $1-\epsilon$,
\begin{align*}
\sup_{f\in\mathcal{F}}\big|\mathrm{E}_{\tau}f-\mathrm{E}^{(T)}f\big|\leq& (1-\tau)D_{\mathcal{F}}(S,T)+2(1-\tau)\mathcal{R}^{(S)}(\mathcal{F})\nonumber\\
&+2\tau\mathcal{R}^{(T)}_{N_T}(\mathcal{F})
+3\tau\sqrt{\frac{(b-a)\ln(4/\epsilon)}{2N_T}}\nonumber\\
&+(1-\tau)
\sqrt{\frac{(b-a)^2\ln(2/\epsilon)}{2}\left(\frac{\tau^2}{N_T}+\frac{(1-\tau)^2}{N_S}\right)}.
\end{align*}
This completes the proof. \hfill$\blacksquare$

\end{document}